\documentclass{article}
\usepackage{arxiv}
\usepackage[textsize=tiny]{todonotes}
\usepackage{comment}
\usepackage{booktabs} 
\usepackage{algorithm}
\usepackage{algpseudocode}
\usepackage{xcolor}
\usepackage{soul}
\usepackage{xspace}
\usepackage{float}
\usepackage{lipsum,graphicx,multicol}
\usepackage{colortbl}
\usepackage{boldline}
\usepackage{amsmath}
\usepackage[export]{adjustbox}
\usepackage[caption=false]{subfig}
\usepackage{pgfplots}
\usepackage{caption}
\usepackage{rotating}
\usepackage[splitrule]{footmisc}
\usetikzlibrary{spy}
\usepackage{setspace}
\makeatletter
\newcommand*{\rom}[1]{\expandafter\@slowromancap\romannumeral #1@}
\newcommand*{\ignore}[1]{}
\makeatother
\usepackage{url}

\title{New Insights into Position Optimization of Wave Energy Converters using Hybrid Local Search}

\author{
  Mehdi Neshat \\
  Optimization and Logistics Group\\
  School of Computer Science\\
  The University of Adelaide\\
   Australia \\
  \texttt{mehdi.neshat@adelaide.edu.au} \\
   \And
 Bradley Alexander \\
  Optimization and Logistics Group\\
  School of Computer Science\\
  The University of Adelaide\\
   Australia \\
  \texttt{bradley.alexander@adelaide.edu.au} \\
  \And
  Nataliia~Y.~Sergiienko \\
  School of Mechanical Engineering\\
  The University of Adelaide\\
   Australia \\
  \texttt{nataliia.sergiienko@adelaide.edu.au} \\
  \And
 Markus Wagner \\
  Optimization and Logistics Group\\
  School of Computer Science\\
  The University of Adelaide\\
   Australia \\
  \texttt{markus.wagner@adelaide.edu.au} \\
  }
\begin{document}
\maketitle

\begin{abstract}
Renewable energy will play a pivotal role in meeting global energy demand in future. Of current renewable sources, wave energy offers enormous potential for growth. This research investigates the optimization of the placement of oscillating buoy-type wave energy converters (WECs). This work explores the design of a wave farm consisting of an array of fully submerged three-tether buoys. In a wave farm, buoy positions strongly determine the farm's output. Optimizing the buoy positions is a challenging research problem due to complex and extensive interactions (constructive and destructive) between buoys.  
This research is focused on maximizing the power output of the farm through the placement of buoys in a size-constrained environment. This paper proposes a new hybrid approach mixing local search, using a surrogate power model, and numerical optimization. 
We compare our hybrid method with other state-of-the-art search methods in five different wave scenarios -- one simplified irregular wave model and four real wave regimes. 
Our methods outperform well-known previous heuristic methods in terms of both quality of achieved solutions and the convergence-rate of search in all tested wave regimes.  
\end{abstract}


\maketitle

\section{Introduction}
Wave energy represents one of the most promising forms of renewable energy due to the high energy density of wave environments and minimal environmental impact \cite{drew2009review}.

One of the current-best designs for wave energy converters (WEC) consists of a large floating buoy tethered to the seafloor.  With this design, energy is produced by the motion of the buoy exerting force on the tether~\cite{mann2007ceto}. 
In some actual deployments, multiple buoys, laid out in a farm, are able to extract power from the waves more than 90\% of the time~\cite{drew2009review}.  
In addition, WECs are able to take advantage of the high energy-density of marine environments -- up to 60~kW per meter of wave front length with a very low impact on aquatic life ~\cite{thorpe1999brief}.
The amount of power produced by a farm or an array of WECs depends on their number, their arrangement with respect to each other, and the prevailing wave conditions \cite{de2014factors}. Thus, generating the appropriate arrangement of WECs in an array is an important factor in maximizing power absorption. 

Currently deployed designs for WECs produce up to 1 MW per buoy~\cite{mann2007ceto}. Thus, to be of commercial scale, it is necessary for farms to consist of multiple buoys. 
However, as the number of converters increases, the optimization of buoy placement becomes more challenging because of the complex hydrodynamic interactions among converters. %
These interactions can be constructive or destructive, and the geometry of these interactions depends strongly on the prevailing wave regime in the environment. 
In evaluating potential layouts, it is important to use an energy-model that has both high-fidelity and simulates the best available WEC designs. The model used in this study \cite{sergiienko2016three} simulates the hydrodynamic behaviour of a fully submerged three-tether WEC in irregular directional waves for several sea sites.

The search space for optimizing array layouts for WECs is multi-modal. Interactions between buoys in an array are complex to model, and the evaluation of each layout is expensive, sometimes taking several minutes. This is because of complex and extensive hydrodynamic interactions between buoys, which in turn depend on the local conditions modelled. These challenges require the use of search meta-heuristics that reliably optimize buoy configurations using a very limited number of layout evaluations.
Work to date on WEC layouts has primarily employed evolutionary algorithms (EAs), which combine stochastic search with selection to progressively improve a population of candidate layouts. In early work \cite{child2010optimal}, Child and Venugopal applied both a simple (and deterministic) Parabolic Intersection (PI) heuristic and a more computationally intensive Genetic Algorithm (GA) to create small (five-buoy) WEC layouts. 

Later work by Sharp and DuPont~\cite{sharp2015wave} used a GA, coupled with heuristics to ensure minimum separation between buoys, to place a small number of WECs (5 converters, 37000 evaluations) in a discretized space. The same authors later explored a similar problem with an improved GA with a cost model \cite{sharp2018wave}.

In both studies, the wave-model used assumed only a single wave direction. The studies also required a relatively large number of layout evaluations, which would limit their application to more detailed wave energy models. 
In \cite{mendes2016particle}, two meta-heuristic algorithms to optimize the geometry of the wave energy generators were introduced, which combines both particle swarm optimization \cite{eberhart1995new} and Box’s complex optimization method \cite{box1965new}. 
An alternative approach was proposed by Ruiz et al. \cite{ruiz2017layout}, who compared the convergence rate and efficiency of three EAs in a discrete search space with a simple wave energy model. The EAs included: CMA-ES \cite{hansen2006cma}, a custom GA, and glow-worm swarm optimization (GSO) \cite{krishnanand2009glowworm}. 
Their work found that search using CMA-ES converged faster than other methods but was outperformed, in terms of energy production by both the GA and GSO. 
In a recent publication, a Differential Evolution with an adaptive mutation operator (IDE) \cite{fang2018optimization} was applied for optimizing a wave farm with three, five and eight converters. Fang et al. proposed some new arrangements of layouts with higher produced energy; however, IDE was not assessed on large wave farms.

In other studies, Wu et al.~\cite{wu2016fast} investigated two popular EAs: the 1+1EA and CMA-ES for optimizing the locations of fully submerged three-tether buoys. The results show that the 1+1EA performed better than  CMA-ES when constrained to a very limited number of layout evaluations. The same wave model was used by Arbones et al.~\cite{arbones2016fast,arbones2018emowave} in a multi-objective optimization problem. In that study, two methods (MO-CMA-ES and SMS-EMOA) were applied to produce good trade-offs between the converter positions, the farm area and required cable length. One of the shortcomings of these approaches, in terms of real-world applicability, is that these works used only a very simple wave model with just one wave frequency and direction. A much more detailed wave scenario was applied in Neshat et al.~\cite{neshat2018detailed}, using an irregular wave model with seven wave directions and fifty sampled wave frequencies, to evaluate a wide variety of EAs and hybrid methods.
This work found that a combination of a stochastic local search combined with the Nelder-Mead simplex method can obtain better 4 and 16-buoy configurations in terms of the total absorbed power.Other aspects of WECs have also been considered. 
For example, Neshat et al.~\cite{mehdi2019layoutAndPTO} considered the optimization of power take-off parameters in addition to the layout optimization for farms to maximize power output. Blanco et al.~\cite{blanco2018dimensioning} used a multi-objective differential evolutionary (DE) solution to optimize operational parameters for a single double-buoy WEC design.

This paper improves on prior work in the following ways: augmenting the findings of \cite{neshat2018detailed} to include another nine new heuristic search methods, including a novel surrogate-based model, all applied to the original irregular wave model; and including four new real wave regimes from the southern coast of Australia (Adelaide, Perth, Tasmania and Sydney) using a higher granularity of wave-directions.
From our experiments, we show that a hybrid framework consisting of a learned model-based local search alternated with numerical optimization outperforms previous heuristic methods in terms of both convergence rate and higher total power output for 16-buoy layouts.

The paper structure is as follows. The next section describes the design of the buoy and the model we use to simulate the intra-buoy interactions. Section~\ref{sec:opt} describes the optimization problem and Section~\ref{Sec:Meta-Heuristic} describes our search methods. We present and discuss our experiments in Section~\ref{Sec:Experimental}. We conclude with a summary and with an outline of possible future work.  
\section{The Numerical Model}
 \subsection{Wave Energy Converter Description}
The wave energy converter considered in this study is shown in Fig.~\ref{fig:wec}. This converter consists of a fully submerged spherical buoy connected to three independent power take-off units through the inclined taut tethers. This design represents a simplified version of the CETO system that is currently under development by Carnegie Clean Energy \cite{mann2007ceto}. 
The parameters of the WEC are specified in Table~\ref{tab:wec}. 

\vspace{3mm}%
\begin{minipage}{\columnwidth}
\small
  \begin{minipage}[b]{0.3\textwidth}
    \centering
    \includegraphics[width=1\textwidth]{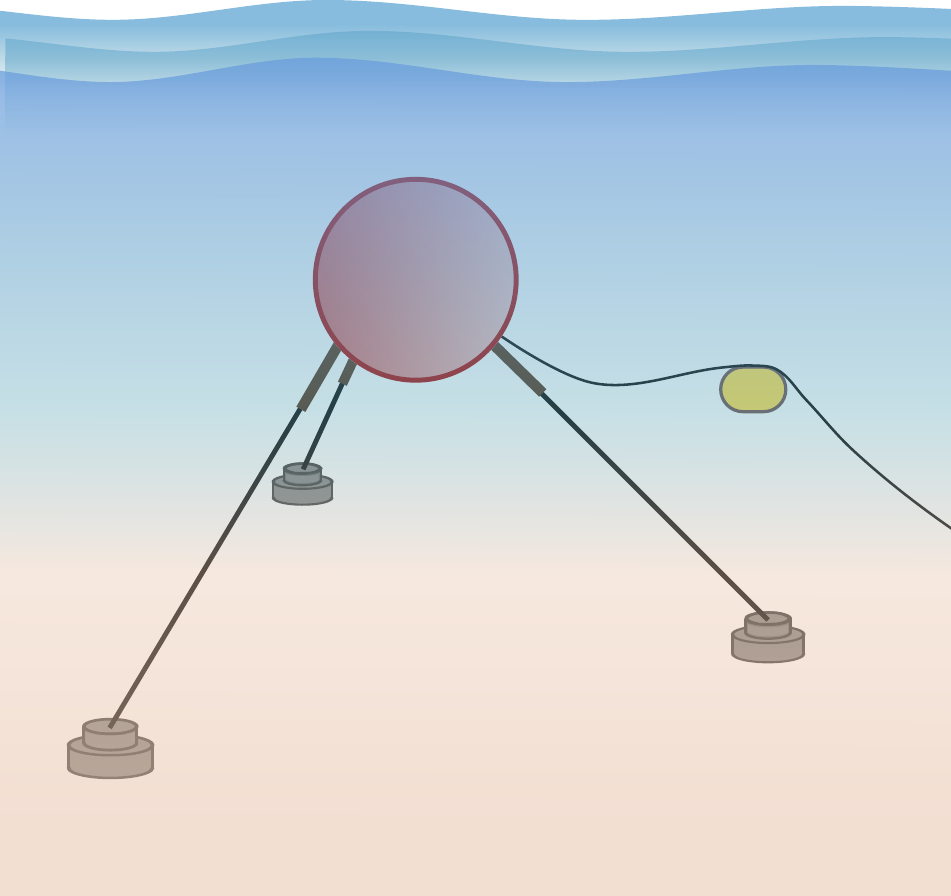}
    \captionof{figure}{WEC}
    \label{fig:wec}
  \end{minipage}
  \hfill
  \begin{minipage}[b]{0.68\textwidth}
    \centering
    \begin{tabular}{lc}\hline
      Parameter & Value \\ \hline
        Buoy radius & 5~m \\
        Water depth & 50~m \\
        Submergence depth & 8~m \\        
        Buoy mass, $m$ & 376$\times 10^3$ kg \\
        Tether angle    & $55^{\circ}$\\
        PTO stiffness, $K_{pto}$ & 2.7$\times 10^5$ N/m \\
        PTO damping, $B_{pto}$ & 1.3$\times 10^5$ Ns/m\\ \hline
      \end{tabular}
      \captionof{table}{Parameters}
      \label{tab:wec}
    \end{minipage}
  \end{minipage}
\vspace{3mm}

The tripod configuration of the power take-off (PTO) system allows power to be absorbed from all three translational degrees of freedom, namely: surge, sway, and heave. As a result, the motion of each WEC can be described by a system of three equations, and a farm consisting of $N$ buoys can be represented in the frequency domain by $3N$ equations assuming linear wave theory:

\begin{equation}
    \left(-(\mathbf{M} + \mathbf{A})\omega^2 + (\mathbf{B} + \mathbf{B}_{pto})j\omega + \mathbf{K}_{pto}\right)\mathbf{x} = 
    \mathbf{F}_{exc}, \label{eq:model}
\end{equation}
Where $\mathbf{M} = m\mathbf{I}_{3N}$ is a mass matrix ($\mathbf{I}_{3N}$ is the identity matrix of size $3N$), $\mathbf{A}$ and $\mathbf{B}$ are the matrices of hydrodynamic added-mass and radiation damping coefficients respectively, $\mathbf{K}_{pto} = K_{pto}\mathbf{I}_{3N}$ and $\mathbf{B}_{pto} = B_{pto}\mathbf{I}_{3N}$ are the PTO stiffness and damping matrices respectively, and $\mathbf{F}_{exc}$ is the frequency dependent vector of excitation forces. All the matrices have dimensions $(3N\times 3N)$. 
The hydrodynamic interaction between submerged spheres, in particular matrices $\mathbf{A}, \mathbf{B}$ and vector $\mathbf{F}_{exc}$, are modeled using a semi-analytical solution presented in \cite{wu1995radiation}. The average power absorbed by all WECs in the farm in a regular wave of unit amplitude, wave frequency $\omega$ and wave angle $\beta$ can be calculated as:
\begin{equation}
    P(\beta, \omega) = \frac{\omega^2}{2}\mathbf{x}^{T}(\beta, \omega)\mathbf{B}_{pto}\mathbf{x}(\beta, \omega), \label{eq:power}
\end{equation}
where $\mathbf{x}(\beta, \omega)$ is obtained solving Eq.~\eqref{eq:model}.

\subsection{Wave Resource}
\begin{figure}[tb]
   \centering
  \subfloat[]{
  \includegraphics[clip,width=0.8\columnwidth]{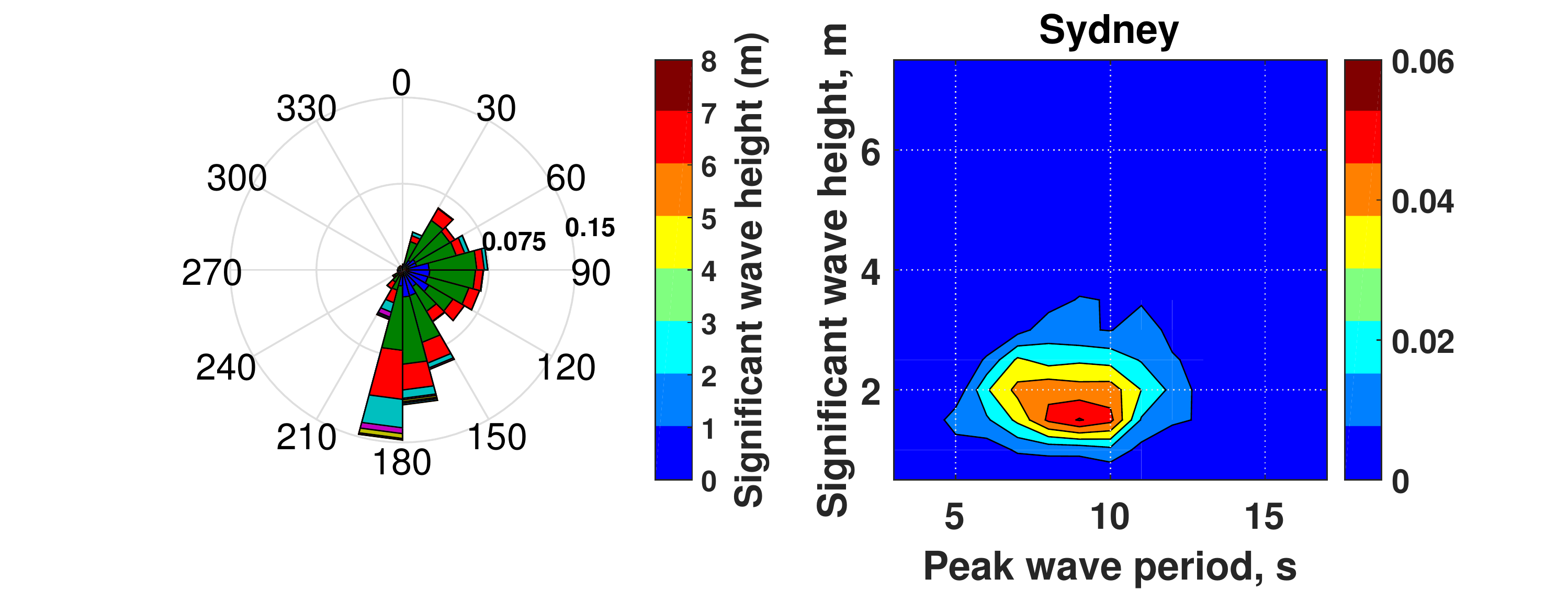}}
  \\
\subfloat[]{
  \includegraphics[clip,width=0.8\columnwidth]{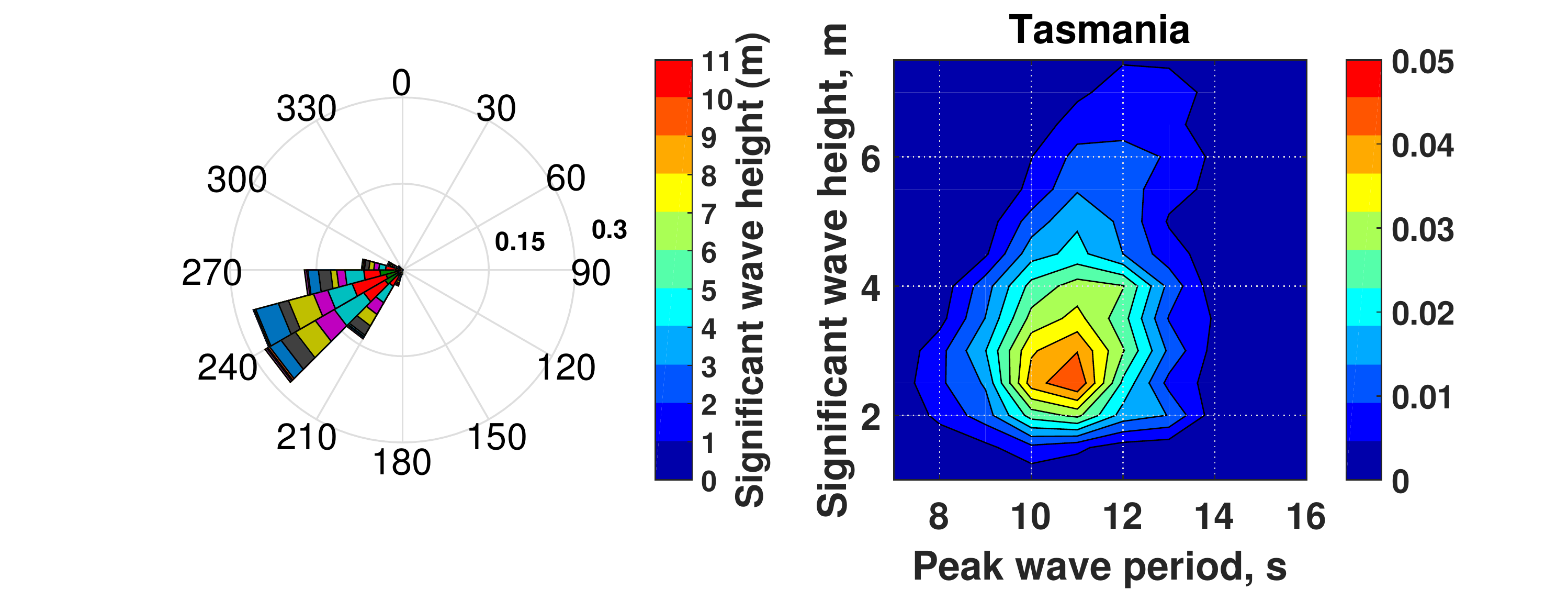} }
\caption{Wave data for two test sites in Australia: (a)~Sydney and (b)~Tasmania. The directional wave rose and wave scatter diagram (left to right).}
\label{fig:wave_direct}
  \end{figure}

We consider four potential sites on the Southern coast of Australia in this study: Adelaide, Perth, Tasmania (southwest coast) and Sydney. The directional wave rose and wave scatter diagram for the Sydney and Tasmania sea sites are shown in Figure.~\ref{fig:wave_direct}. 
These underlying wave data was obtained from the Australian Wave Energy Atlas~\cite{CSIRO2016}. 

\subsection{Wave Farm Performance Evaluation}
The total average annual power ($P_{AAP}$) produced by the wave farm is calculated by summing the contribution of energy absorption from each of the sea states representing a wave climate \cite{flavia2017numerical}:
\begin{equation}
P_{AAP} = \sum_{i=1}^{N_s} O_i(H_s, T_p)P_i(H_s, T_p), \label{eq:real_wave}
\end{equation}
Where $N_s$ is a number of sea states considered, $H_s$ is the significant wave height, $T_p$ is the peak wave period of the sea state, $O_i(H_s, T_p)$ is the probability of occurrence of a sea state (which can be derived from the wave scatter diagram) and $P_i(H_s, T_p)$ is the power generated by the wave farm in the $i$-th sea state. A sea state refers to the condition of the sea/ocean surface that can be characterized by statistics (significant wave height and peak wave period).

For irregular waves, $P_i(H_s, T_p)$ is the sum of power contribution from each frequency in the spectrum and each wave direction \cite{flavia2017numerical}:
\begin{equation}
P_i(H_s, T_p) = \int_0^{2\pi}\int_0^{\infty}2S_i(\omega)D(\beta)P(\beta,\omega)\mathrm{d}\omega \mathrm{d}\beta, \label{eq:simplified}
\end{equation}
Where $P(\beta,\omega)$ is calculated according to Eq.~\eqref{eq:power}, $S_i(\omega)$ is the irregular wave spectrum (the Bretschneider spectrum considered in this study) and $D(\beta)$ is the directional spreading spectrum specific for the site (obtained from the directional wave rose).

The hydrodynamic interaction among converters in the wave array affects its power production and is usually quantified using the $q$-factor, defined as:
\begin{equation}
q = \frac{P_{AAP}}{NP^{isolated}_{AAP}},
\end{equation}
where $N$ is a number of WECs forming the array, $P^{isolated}_{AAP}$ is the power generated by an isolated WEC. If the wave interaction has, on average, a constructive effect on the power production of the array, then $q>1$, and if the effect is destructive then, $q<1$.

In this work, all optimization algorithms are, first, evaluated using a simplified synthetic wave model that corresponds to the most frequently occurring sea state at the Sydney site ($H_s = 2$ m, $T_p = 9$ sec) using Eq.~\eqref{eq:simplified}. Subsequently, the algorithms are tested for four 
real wave scenarios using Eq.~\eqref{eq:real_wave}.

\section{Optimization Setup}\label{sec:opt}
Using the wave model, the optimization problem can be stated in terms of positioning $N$ WECs over a bounded area of a wave farm $\Omega$ in order to maximise the total power production $ P_{AAP}$.

\begin{equation}
     \begin{aligned}
      \text{} & P_{AAP}^*= \mbox{\em argmax}_{\mathbf{x,y}} P_{AAP}(\mathbf{x_i,y_i})  \\
      \text{Subject to} &\\
       & [x_i,y_i]\in \Omega, \;\; i=1,...,N \\
       & \mbox{\em dist}((x_i,y_i),(x_j,y_j))\ge R^\prime\;\;i\neq j=1,...,N 
     \end{aligned}
\end{equation}
where $P_{AAP}(\mathbf{x,y})$ is the sum of mean power output by buoys positioned in an area at $x$-positions $\mathbf{x}=[x_1,\ldots,x_N]$ and corresponding $y$ positions $\mathbf{y}=[y_1,\ldots,y_N]$. In this study, the maximum number of buoys is predefined to be  $N=16$. Each buoy $i$'s position is expressed as the coordinate: $[x_i,y_i]$. This position is constrained to be within the area $\Omega$. Where $\Omega=l\times w $ and $l=w=\sqrt{N * 20000}\,m$. This constraint is given to model the scenario where there are strict limits on the area allotted to a wave farm lease. A second constraint is a minimum separation between buoys ($R^\prime=50m$), representing a gap required for maintenance vessels to safely pass. 
For each array, $\mathbf{x,y}$ the sum-total of the safety distance violations is: 

$\mbox{\em{Sum}}_{\mbox{\em dist}}= \sum_{i=1}^{N-1}\sum_{j=i+1}^{N} 
(\mbox{\em{dist}}((x_i,y_i),(x_j,y_j))-R^\prime, $\\
\hspace*{30mm}$\mbox{if } \mbox{\em{dist}}((x_i,y_i),(x_j,y_j))<R^\prime$ \mbox{else 0}

\noindent where $\mbox{{\em dist}}((x_i,y_i),(x_j,y_j))$ is the Euclidean distance between buoys $i$ and $j$. 
To provide a smooth response to such violations, we apply a steep penalty function $(\mbox{\em{Sum}}_{\mbox{\em{dist}}}+1)^{20}$ to the total power output (in Watts). 
\subsection{Computational Resources}

In this work, to compare methods fairly we allocate a uniform time budget for each optimization run of three days on the dedicated platform with a 2.4GHz Intel 6148 processor running 12 processes in parallel with 128GB of RAM. The software environment running the function evaluations and the search algorithm is MATLAB R2016.
In order to maximize the use of the time budget the algorithms are, depending on the search heuristic, written to evaluate the energy in parallel either per wave frequency, or per-individual. This configuration achieves an approximately 10-fold speedup for each algorithm tested. 
Exceptions to this principle of making maximum use of the time budget are made for search methods that converge quickly and produce little pay-off for additional evaluations.  
To allow for a valid statistical comparison we repeat all search methods  10 times. 

\section{Meta-Heuristic Search Techniques}
\label{Sec:Meta-Heuristic}
Meta-heuristic optimization methods have been applied extensively in fields where the global search is needed, including in the field of renewable energy technologies optimization \cite{dubey2018overview}. 
This paper compares the methods described in our previous work~\cite{neshat2018detailed} to nine new algorithm variants derived for this work, and to other recent approaches. This previous work compared the performance of random search (R-S); forms of partial evaluation (PE)~\cite{neshat2018detailed} where layouts are evaluated on random subsets of frequencies; TDA~\cite{wagner2013fast}, used for wind-farm layout; CMA-ES~\cite{hansen2006cma}; Differential Evolution (DE)~\cite{storn1997differential}; Improved Differential Evolution \cite{fang2018optimization}; binary Genetic Algorithm \cite{sharp2018wave}; 1+1EA's with various mutation settings; local or neighbourhood search (LS); and Nelder-Mead downhill search (NM). In this earlier work the best performing heuristic, $LS_3-NM_{2D}$ combined one-at-a-time buoy placement with iterated local search and Nelder-Mead to refine each buoy position. The algorithms described here improve significantly on the performance of this earlier work by exploiting knowledge specific to the target wave scenario. These new (smart) search heuristics are described next. 
All of the heuristic methods that are compared in this paper are listed in Table~\ref{table:details:OP}.

\subsection{Smart Local Search (SLS)}
\begin{figure*}[tbp]
  \includegraphics[width=\textwidth]{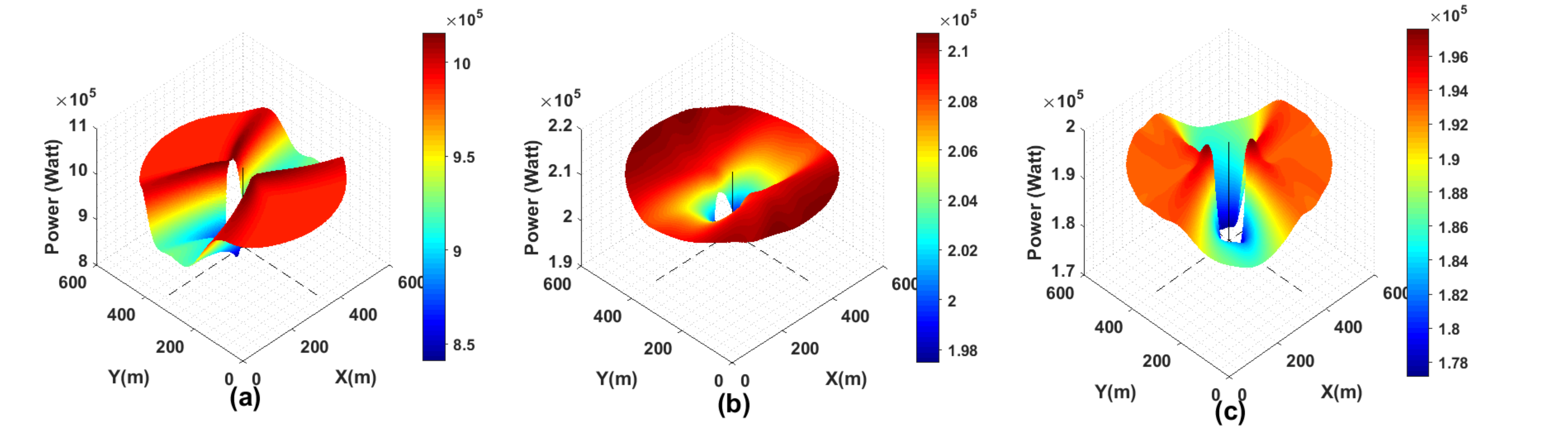}
   \caption{The 3D power landscape a two-buoy array based on the simplified irregular (a), Sydney (b) and Perth (c) wave scenarios. The first buoy's position is fixed,  the second buoys is  varied to measure total energy output. The mapped area extends $360 \circ$ and a distance of between 50(m) and 300(m). 
   }
   \label{fig:3D_Power_position}
 
  \end{figure*}

\begin{figure*}[htbp!]
  \includegraphics[width=\textwidth]%
  {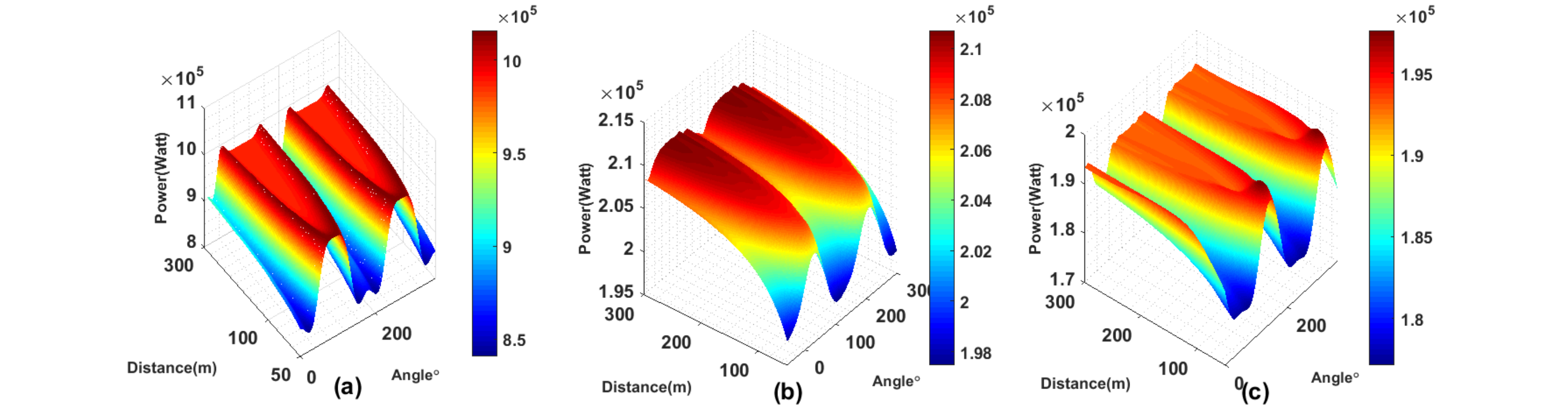}
  \caption{A 3D power landscape, for relative angle and distance between two buoys  based on the simplified irregular wave model (a) and two real wave scenarios: Sydney (b) and Perth (c). 
  Note that there are ridges in the power landscape corresponding to areas of constructive interference. The Improved Smart Local Search algorithm variants, described in this paper, exploit this local landscape when placing buoys. }
   \label{fig:power_angle_distance}
  \end{figure*}

\begin{figure}[tbp]
\centering\includegraphics[width=0.6\textwidth]{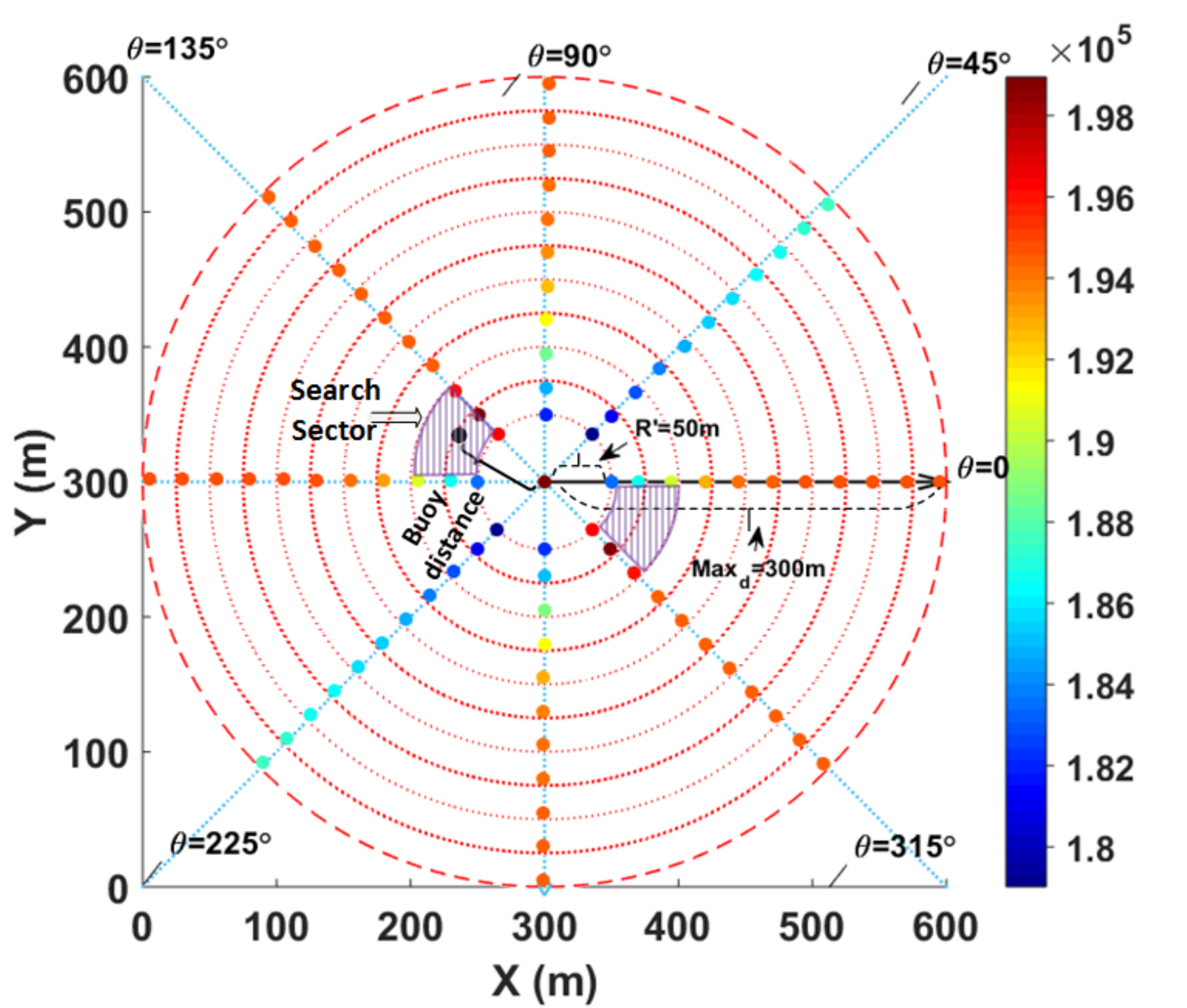}
   \caption{The 2D power landscape of two-buoy array based on the Adelaide wave scenario. 
   }
   \label{fig:SLSPatterns}
  
  \end{figure}

 In previous work, we have observed that a good candidate position for the next buoy is in the neighborhood of the previous buoy. The SLS method improves upon these earlier searches by placing the next buoy in a relative position informed by peaks in the power landscape built from sampling positions in a two-buoy model under local wave conditions. 
Examples of such landscapes are given in Figure.~\ref{fig:3D_Power_position}, which shows a 3D power landscape of the simplified irregular, Sydney, and Perth wave models. It can be seen from these landscapes that, even for two buoys, there is variation in the shape of the landscape and the positions of the point at which there are constructive interactions. Note that, for a given wave regime, it is not practical to infer the shape of this power landscape by means other than sampling it. Furthermore, the inter-relationship among the absorbed power, angle and distance between two-buoy layout for different wave scenarios can be seen in Figure \ref{fig:power_angle_distance}. 
In the SLS method, this two-buoy power landscape is sampled. The pattern of sampling into this landscape is shown in Figure.~\ref{fig:SLSPatterns}. This sampling landscape has an angular resolution of 45-degree intervals and a distance resolution of 5m intervals. This sampled landscape is computationally cheap to build because it models interactions between only two buoys. Moreover, this sampling exercise only has to be run once for each wave scenario at the beginning of the search process. These samples are then used to define the most promising sectors, called the {\em{search-sectors}}, in the search landscape for the placement of the next buoy. These sectors, marked in Figure.~\ref{fig:SLSPatterns} with a dotted line, lie between the best and second-best points in the search landscape on either side of the current buoy.
 
 \begin{center}
 \scalebox{0.99}{
    \begin{minipage}{\linewidth}
\begin{algorithm}[H]
\caption{$\mathit{Smart Local Search}$}
\label{alg:SLS}
\begin{algorithmic}[1]
\Procedure{Smart Local Search}{}\\
 \textbf{Initialization}
 \State Generate surrogate 2-buoy power model
\State $\mathit{size}=\Omega$  \Comment{Farm size}
\State $\mathit{ pos}=[(x_1,y_1),\ldots,(x_N,y_N)]=\bot$\Comment{positions}
\State $\mathit{pos}(1)=(\mathit{size}/2,0)$ \\
 \textbf{Search}
\For{ $i$ in $[1,..,N]$ }
  \State update search sectors $S$
  \State $\mathit{bestEnergy}=0$
  \State $\mathit{bestPosition}=(0,0)$
  \For{ $j$ in $[1,..15]$} \Comment{$15$ random samples}
      \State $(x_s,y_s)=U(S)$ \Comment{sample sectors}
      \State $\mathit{pos}(i)=(x_s,y_s)$
      \State $\mathit{energy}=\mathit{Eval}(\mathbf{pos})$
      \If{$\mathit{energy}>\mathit{bestEnergy}$}
      	\State $\mathit{bestEnergy}=\mathit{energy}$ 
        \State $\mathit{bestPosition}=(x_s,y_s)$
      \EndIf
  \EndFor
  \State $\mathit{pos}(i)=\mathit{bestPosition}$
  \EndFor
  
\State \textbf{return} $\mathit{pos}$ \Comment{Final Layout}
\EndProcedure
\end{algorithmic}
\end{algorithm}
 \end{minipage}%
     }
     \end{center}

 \begin{table} 
 \small
\caption{Summary of the search methods used in this paper. All methods are given the same computational budget. Parallelism can be expressed as per-individual or per-frequency depending on the number of individuals in the population from section.\ref{Sec:Meta-Heuristic}.}
\centering
\label{table:details:OP}
\scalebox{0.9}{
\begin{tabular}{|l|l|p{10.5cm}|}
\hline 
 
Abbreviation & Parallelism & Description\\ \hline\hline
\normalsize{R-S}  & per-frequency & Random Search \\ 
\hline 
\normalsize{PE$_{50\: \mu}$} &  per-individual & Partial Evaluation\cite{dang2016runtime}, all frequencies (PE-FULL), population $\mu\in\{10,50,100\}$ \\
\hline
PE$_{f\: \mu}$ & per-individual & Partial Evaluation \cite{dang2016runtime}, partial frequencies, $f\in\{1,4,16\}$, $\mu\in\{10,50,100\}$ \\ 
\hline
TDA &  per-individual& Algorithm for optimizing wind turbine placement from~\cite{wagner2013fast} \\ 
\hline
CMA-ES & per-individual  & CMA-ES\cite{hansen2006cma} all dimensions, $\mu='4 + int(3 * log(D)){\mbox{ndim}}$ , $\sigma=0.17*size$ \\ 
\hline
CMA-E$S_{PF}$ & per-frequency  & All settings are like CMA-ES \\ \hline
CMA-ES (2+2) & per-individual & Setup for CMA-ES from~\cite{wu2016fast}, $\sigma=20m$ \\
\hline
CMA-E$S_{PF}$ (2+2) & per-frequency & All settings are based on ~\cite{wu2016fast} \\ 
\hline
DE$_{\mbox{$P_{cr}$}}$ & per-individual & Differential evolution \cite{storn1997differential}, $\mu=50$, $F=0.5$, $\mbox{$P_{cr}$}\in \{0.3,0.5,0.7,0.9\}$ \\
\hline
Improved DE & per-individual & Improved Differential evolution , All settings are based on \cite{fang2018optimization}\\ 
\hline
bGA & per-individual & binary Genetic Algorithm, All settings are based on \cite{sharp2018wave}\\ 
\hline
1+1EA$_\sigma$ & per-frequency & 1+1EA(all dimensions), mutation step size with $\sigma\in {3,10,30}$(m) \\
\hline
1+1EA$_s$ & per-frequency & 1+1EA (all dimensions)  with uniform mutation in range $[0,s]$ with $s=30m$ from \cite{wu2016fast} \\ \hline
1+1EA$_{\mbox{Linear}}$ & per-frequency & 1+1EA (all dimensions) with linearly decaying mutation step size \cite{neshat2018detailed}\\ 
\hline
1+1EA$_{1/5}$ & per-frequency & 1+1EA (all dimensions) with adaptive step size \cite{eiben1999parameter} \\ \hline
Fuzzy-1+1EA & per-frequency & 1+1EA (all dimensions) with fuzzy adaptive mutation step size \\
\hline
Iterative-1+1EA & per-frequency & Iterative local search - buoys are placed in sequence using best of local neighborhood search \cite{neshat2018detailed} \\
\hline
LS-NM$_{\mbox{allDims}}$ & per-frequency & Local search + Nelder-Mead search in all Dimensions \cite{neshat2018detailed} \\ \hline
NM\_Norm$_{2D}$ & per-frequency & Buoys placed in sequence using Nelder-Mead search, Initial placement normally distributed from last buoy position \cite{neshat2018detailed}\\ 
\hline
NM\_Unif$_{2D}$ & per-frequency & Buoys placed randomly and then refined using Nelder-Mead Initial placement uniformly distributed from last buoy position \cite{neshat2018detailed} \\
\hline
L$S_{1}$-N$M_{2D}$  & per-frequency & Local Sampling + Nelder Mead search. Buoys placed at random offset from previous buoy and placement refined by Nelder-Mead search. \cite{neshat2018detailed} \\
\hline
L$S_{3}$-N$M_{2D}$ & per-frequency & Iterative local search + Nelder-Mead search. Placements sampled at 3 random offsets from previous location, best placement used as starting point for Nelder-Mead search. \cite{neshat2018detailed} \\
\hline
SLS &  per-frequency & Providing the two-buoy power landscape+ Extracting a proper domain of the distances and the angles +Iterative local search +Smart Mutation;Uniform distribution, 15 samples, step= $rand$($R'$, Buo$y_{Distance}+\kappa_2(20m)$). 4.5 folds faster than the best method of the prior study \cite{neshat2018detailed} for 16-buoy layout.  \\
\hline
SLS-NM &  per-frequency &  Smart Local Search with three samples of the mutation+ Nelder-Mead search\\ 
\hline
ISLS &  per-frequency &   Improved Smart Local Search : Creating a more accurate knowledge-based surrogate power model, placing a new buoy: the initial sequential $N_{sb}$-buoy number $\sigma=R^\prime$ and for next buoys $ \sigma=2\times R^\prime$: Mutating:10 samples for initial sequential $N_{sb}$-buoy number, and for next buoys 20 samples, $step= rand(R^\prime,$Buo$y_{Distance}$+10m). 60\% faster than L$S_3$-N$M_{2D}$in  \cite{neshat2018detailed} for 16-buoy layout.  \\
\hline
ISLS-NM & per-frequency & Improved Smart Local Search (10 samples) for the initial sequential $N_{sb}$-buoy number and for last buoys 3 samples + Nelder-Mead search,     \\
\hline
ISL$S_{(II)}$-F&  per-frequency & Improved Smart Local Search (for initial sequential $N_{sb}$-buoy number) (3 samples)+  Applying SQP (for finding the furthest point of the area based on the layout position).  20 times faster than the best method of the prior work \cite{neshat2018detailed} for 16-buoy layout.      \\ 
\hline
ISL$S_{(II)}$-NM &  per-frequency & Improved Smart Local Search(II) (for initial sequential $N_{sb}$-buoy number) 3 samples + Nelder-Mead Search. \\
\hline
ISL$S_{(II)}$-SQP &  per-frequency &Improved Smart Local Search(II) (for initial sequential $N_{sb}$-buoy number) 3 samples + Sequential Quadratic Programming Search.  \\ 
\hline
ISL$S_{(II)}$-AS& per-frequency & Improved Smart Local Search(II) (for initial sequential $N_{sb}$-buoy number) 3 samples + Active-Set Search.  \\
\hline
ISL$S_{(II)}$-IP &  per-frequency &Improved Smart Local Search(II) (for initial sequential $N_{sb}$-buoy number) 3 samples + Interior-Point Search.  \\
\hline
\end{tabular}
}

\end{table}

For the placement of each buoy, a local search makes 15 samples, uniformly distributed in the search sectors (subject to boundary constraints). These samples are assessed with the full model, which calculates {\em{all}} buoy interactions in the current array. From these 15 samples, the best buoy location is selected. 

Analysis of our experiments has shown that 15 samples have been sufficient to improve upon the initial placement with a probability of $99\%$ with an expected improvement of power production almost identical to that of a much larger number of samples \cite{neshat2018detailed}. Algorithm~\ref{alg:SLS} describes the SLS method. 


\subsection{Smart Local Search + Nelder-Mead (SLS-NM)}

 Smart Local Search + Nelder-Mead (SLS-NM) explores the same search sectors as the SLS algorithm defined above. The SLS-NM algorithm differs in that it takes only three random samples from the search sectors and uses the best of these as the start point for a Nelder-Mead (NM) simplex search process. The NM search process can robustly move to a local optimum from its starting point. In order to fit within the computational budget, the number of steps (evaluations) in the NM search is limited to a maximum of $20$. To verify that this number of steps is adequate we ran a longer experiment with $120$ evaluations for each NM search, and we found that, in most cases, the improvement from the extra evaluations was less than 2\% of the power output for that buoy. Moreover, when there was a significant improvement, it happened only after a prohibitively large number of evaluations.

\subsection{Improved Smart Local Search (ISLS)}
From our analysis of experimental runs using the SLS methods, we made four observations.
\begin{enumerate}
\item The search sector containing the best-sampled positions was always in the direction of the opposite side of the farm.
\item The best samples from the SLS-NM method came from a sector which was narrower in angular extent but longer in radial extent. 
\item The search space for placing the next buoy becomes much harder after hitting the top boundary of the farm. This is due to occlusion from the front row of buoys for subsequent buoy placement. 
\item The placement of the first buoy in the centre of the bottom boundary of the farm can be sub-optimal if the best angle for the alignment of buoys in the surrogate power landscape causes some buoys to encounter the left or right boundary of the farm. 
\end{enumerate}

In response to these observations, we designed a refined search method called Improved Smart Local Search (ISLS). This search method addresses the first observation above by only sampling the search sector in the direction of the current opposite boundary of the farm (upwards in our implementation).  ISLS addresses the second observation by allowing the user to set the angular and radial extent of the search sector for the wave scenario. The third observation is addressed by reducing the number of samples used when placing buoys on the first sweep to the opposite farm boundary (phase 1) and running more samples to place subsequent buoys (phase 2). Lastly, the fourth observation is addressed by placing the first buoy in the left corner of the landscape if the best angle is between zero and 90 degrees and in the right corner otherwise. 

\begin{figure}[tbhp]
\centering\includegraphics[width=0.5\textwidth,height=11cm]{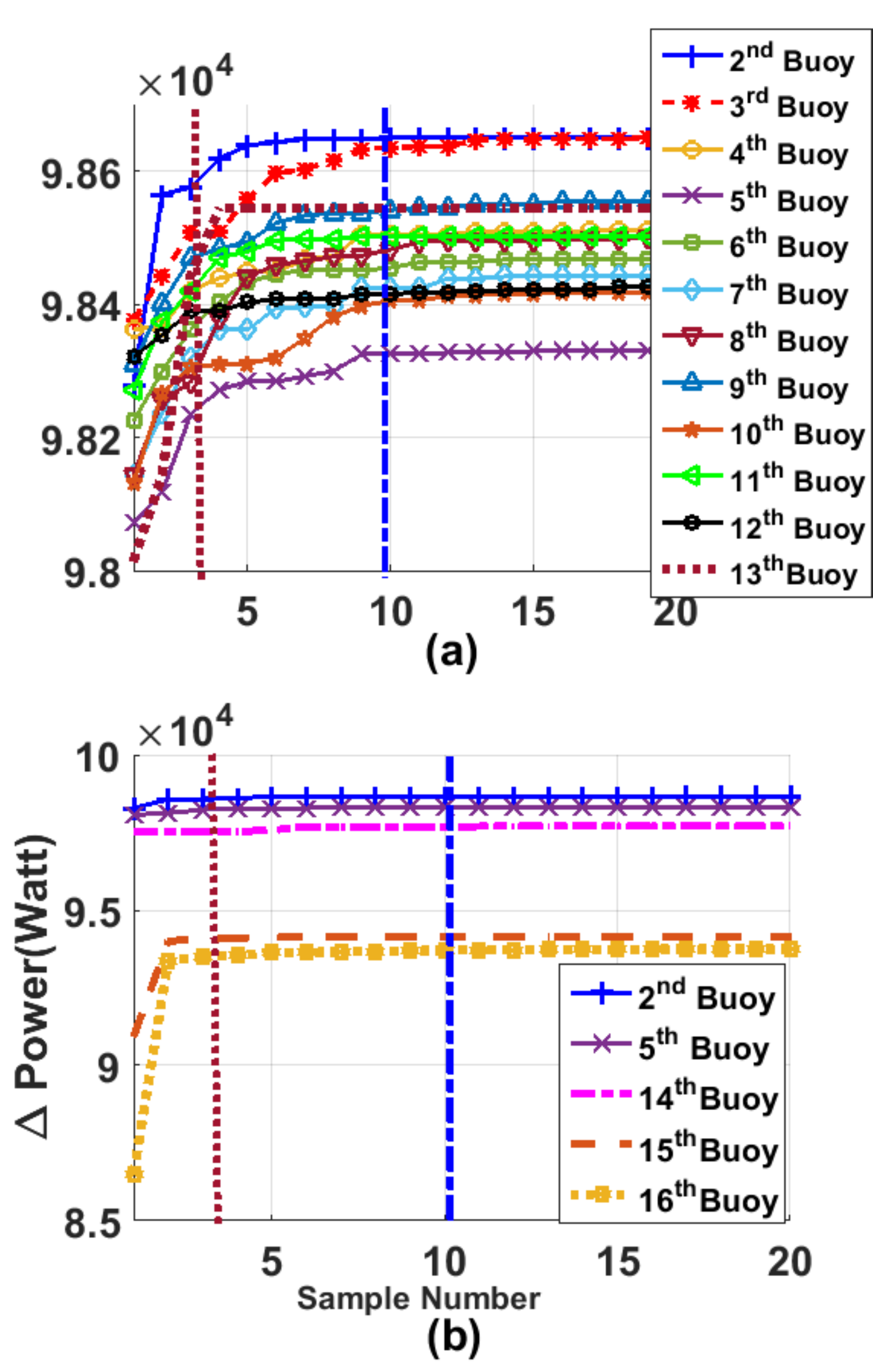}
   \caption{The impact of sample number on the ISLS performance (16-buoy) for the Perth wave model. The red vertical line shows 3-sample.}
   \label{fig:ISLS_sampling}
\end{figure}
We base the decision on the number of samples to use in the first sweep of buoy placement (phase 1) and then for the placement of subsequent buoys (phase 2) on empirical studies on the impact of different numbers of samples depending on the different wave scenarios. 

To illustrate the findings of this process, Figure~\ref{fig:ISLS_sampling}(a) shows the average power gain from sampling for the first 12 buoys for ten runs in the Perth wave scenario. The red and blue vertical lines, respectively, indicate the average power gained after three and ten samples. For the experiments leading to this Figure, the buoys were placed according to the best result obtained after 20 samples. The Figure indicates that for the placement of the first 12 buoys (phase 1) the curves flatten after ten samples, with buoy three showing the largest gain, after ten samples, of $0.018\%$. This indicates that we gain little from sampling the real power landscape beyond this point. Thus for phase 1, 10 samples were allocated for each buoy placement. For the placement of the last four buoys (phase 2), as illustrated in Figure ~\ref{fig:ISLS_sampling}(b) the gain from sampling is steeper and, for buoy 16, the improvement between $10$ and $20$ samples is $0.032\%$. 

A second notable feature of Figure~\ref{fig:ISLS_sampling}(a) is that the power curves for some buoys are vertically displaced relative to others. This indicates that for some buoy placements the power landscape is more challenging. In general, we have observed that the displacement of these sampling curves for later buoys depends on the placement of previous buoys. In some experiments, we have observed that this dependence on placement history can even lead to some minor anomalies in search behaviour where sampling {\em{less}} for earlier buoy placements appears to make the search landscape slightly easier for subsequent buoy placements. However, we have observed for all wave regimes that the best {\em{median}} performance for wave farms is obtained by employing as many samples as the time budget allows in both phases of the search process. 

\subsection{Improved Smart Local Search Nelder-Mead (ISLS-NM)}
As previously noted, due to occlusion by other buoys, the power landscape for phase~2 of the search is different from that for phase~1. This means that, for phase~2 buoy placements, the search sector from the surrogate power landscape might not contain the best location for the placement of the next buoy. To search more broadly a variant of ISLS, called ISLS-Nelder-Mead (ISLS-NM) was created. This variant has an identical phase~1 search to ISLS, but in phase~2 it performs a local search with three samples followed by 20 samples of Nelder-Mead search starting at the best of these three sampled locations. In almost all experimental runs, these 20 samples were enough to converge to a point where the step size is less than $1\%$ of the total power output for that buoy. 

\subsection{Improved Smart Local Search-II (ISLS-II)}
One drawback of ISLS is the need for the user to define the angular and radial extent of the search sector for a wave regime through observation of the surrogate power landscape. In ISLS$_{II}$ this process is automated by, first performing fine-grained sampling of the 2-buoy power landscape. The readings from this sampling are arranged into a table -- with columns representing angular increments and rows representing radial distance increments. The maximum power value is then located. The search sector is then defined by the area between this highest and second-highest sample -- subject to a maximum radial distance constraint of 300m. As with ISLS, this method tends to produce a longer and narrower search sector than SLS.

We have implemented five variants of ISLS$_{II}$. Apart from the determination of the search sector, each of these variants has an identical first phase to ISLS using the ten-sample randomized local search within the search sector that is defined by a surrogate landscape. 
Each of these variants, in their second phase, still begins by identifying the best of three random sample positions in the search sector relative to the previously placed buoy. The variants differ, however, in the type of search that proceeds from each sample point. In the following, we  describe these search variants in turn. 
\paragraph{ISLS-II + Active Set (ISLS$_{II}$-AS)}
Here, we refine the placement of each buoy with 20 evaluations of the Active-Set \cite{nocedal2006nonlinear} search method. This method identifies the set of boundary constraints that are relevant to the current state of search and concentrates search close to these. This method is able to take large steps through the search space, thus allowing for quick coverage of the search area. This variant is described in Algorithm \ref{alg:ISLS(II)-AS}.

 \begin{center}
 \scalebox{0.95}{
    \begin{minipage}{\linewidth}
    \begin{algorithm}[H]
\caption{$\mathit{ISLS(II)-AS}$}\label{alg:ISLS(II)-AS}
\begin{algorithmic}[1]
\Procedure{Improved Smart Local Search(II) + Active-Set}{}\\
 \textbf{Initialization}
\State Generate surrogate 2-buoy power model
\State $\mathit{size}=\Omega$  \Comment{Farm size}
\State $\mathit{SafeDis}=50$  \Comment{safe distance between buoys}
\State $\mathit{pos}=[(x_1,y_1),\ldots,(x_N,y_N)]=\bot$\Comment{positions}

\If {$0^\circ<\mathit{bestLocalAngle}<90^\circ$}
	 $\mathit{pos}(1)=(0,0)$ 
\Else
     $~\mathit{pos}(1)=(\mathit{size},0)$
\EndIf 
\State $\mathit{BN_{row}}=round(size/\cos(bestLocalAngle))$  \Comment{buoy number in first row}
\State $\mathit{buoyNum}=2$\\
\textbf{Search}
\State update search sector $S$
\While{$\mathit{bottomYBoundary}(S)<\mathit{size}$}\Comment{phase 1}
  \State $\mathit{bestEnergy}=0$
   $,\mathit{bestPosition}=(0,0)$
  \For{ $j$ in $[1,..,10]$} \Comment{$10$ random samples}
      \State $(x_s,y_s)=U(S)$ \Comment{sample sector}
      \State $\mathit{pos}(i)=(x_s,y_s)$
      \State $\mathit{energy}=\mathit{Eval}(\mathbf{pos})$
      \If{$\mathit{energy}>\mathit{bestEnergy}$}
      	\State $\mathit{bestEnergy}=\mathit{energy}$ 
        \State $\mathit{bestPosition}=(x_s,y_s)$
      \EndIf
   \EndFor
   \State $\mathit{pos}(i)=\mathit{bestPosition}$
   \State $\mathit{buoyNum}=\mathit{buoyNum}+1$
   \State update search sector $S$
\EndWhile
\For{ $i$ in $[\mathit{buoyNum},..,N]$ } \Comment{phase 2}
  \State $\mathit{bestEnergy}=0$
   $,\mathit{bestPosition}=(0,0)$
  \For{ $j$ in $[1,..3]$} \Comment{$3$ random samples}
      \State $(x_s,y_s)=U(S)$ \Comment{sample sector}
      \State $\mathit{pos}(i)=(x_s,y_s)$
      \State $\mathit{energy}=\mathit{Eval}(\mathbf{pos})$
      \If{$\mathit{energy}>\mathit{bestEnergy}$}
      	\State $\mathit{bestEnergy}=\mathit{energy}$ 
        \State $\mathit{bestPosition}=(x_s,y_s)$
      \EndIf
  \EndFor
   \If{$(\mathit{bouyNum}\le \mathit{BN_{row}})\lor(\mathit{bestPosition}\ge \mathit{size-SafeDis} )$}
        \State $\mathit{bestPosition}=\mathit{ActiveSearch}(\mathit{bestPosition},20)$
  \State $\mathit{pos}(i)=\mathit{bestPosition}$
   \EndIf
  \State update search sector $S$
  \EndFor
  
\State \textbf{return} $\mathit{pos}$ \Comment{Final Layout}
\EndProcedure
\end{algorithmic}
\end{algorithm}
\end{minipage}%
     }
     \end{center}
   \paragraph{ISLS-II +  Sequential Quadratic Programming(ISLS$_{II}$-SQP)}

This approach refines the placement of each buoy by performing Sequential Quadratic Programming (SQP)~\cite{xiao2014optimal}. This search method employs Newton's method when the search is away from boundary constraints and reverts to constrained search when boundaries are encountered.
\paragraph{ISLS-II + Fast placement (ISLS$_{II}$-F)}
We observed in the earlier ISLS-NM that the phase~2 buoy placements tended to reside on the lee-side behind the front row of buoys, with these buoys finishing far from each other. Informed by this observation the ISLS$_{II}$-F algorithm uses 20 iterations of SQP search to place each buoy, one at a time, at a position that is the maximum Euclidean distance from the previously placed buoys. Note that using distance as a proxy function makes this method very fast compared to other variants.


\paragraph{ISLS-II + Nelder-Mead (ISLS$_{II}$-NM)}
This approach is similar to the earlier ISLS-NM, but it applies 20 iterations of the Nelder-Mead algorithm to the placement of each buoy.

\paragraph{ISLS-II + Interior point algorithm (ISLS$_{II}$-IP)}
This algorithm refines each buoy position using the interior-point (IP) algorithm \cite{glavic2004interior} for constrained search. This method is similar to other active-set methods above except that boundaries are approximated using barrier functions which allow search near constraint boundaries rather than on constraint boundaries.  
This concludes our description of the different buoy placement algorithms explored in this paper. The next section presents detailed results comparing the performance of these algorithms.

\section{Experimental Results }
\label{Sec:Experimental}
This section presents the results of the experiments comparing the performance of the algorithms described above on the placement of buoys under the different wave scenarios. 

\begin{figure}[htbp]%
\includegraphics[width=\textwidth]{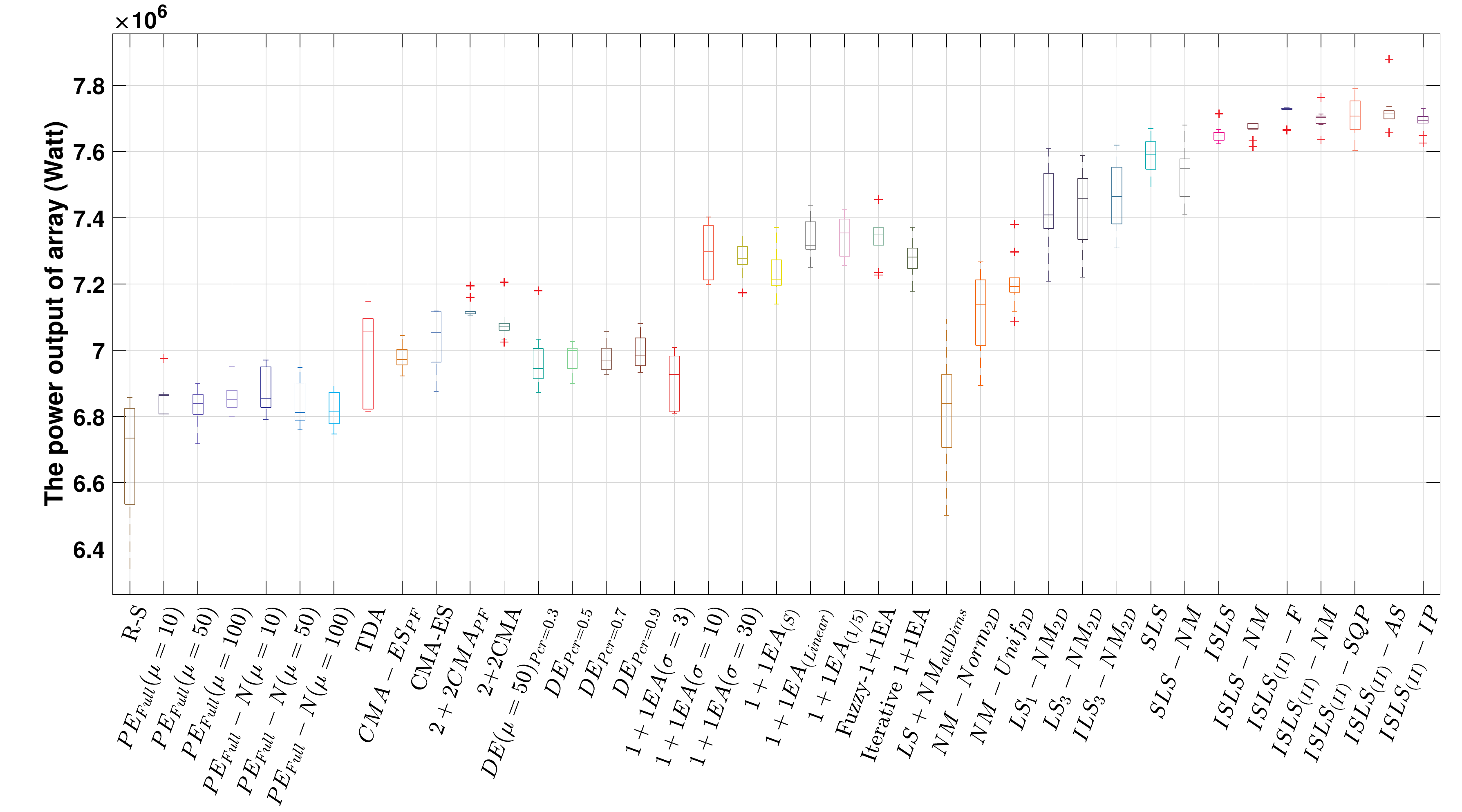}
\caption{The performance comparison of the best layout per run for the proposed heuristic approaches for optimizing the position of 16 buoys under the simplified wave scenario. ISLS$_{(II)}$-AS produces $\textbf{4\%}$ more power compared with the best result in~\cite{neshat2018detailed}.} 
 \label{fig:all_comparion_regular}
 \end{figure}

One challenge for the approaches is that the farm's dimensions do not allow for all 16 buoys to be placed in a single line. Another challenge is that, because of interactions, the cost of the evaluative model scales quadratically with the number of buoys. This means the number of full evaluations of a 16-buoy layout within our 3-day time budget is limited to a few thousand evaluations. This limited computational budget heavily favours problem-specific search algorithms that place one buoy at a time. This is illustrated for the simplified irregular wave scenario in Figure.~\ref{fig:all_comparion_regular}. This box-and-whiskers plot demonstrates the best-achieved power for a 16-buoy layout of each search framework. 
The new methods described in this paper are given in the last nine columns of the Figure. The improved performance of these new search methods is quantified in Table~\ref{tab:all}. 
\begin{sidewaystable}
\centering
\setlength{\tabcolsep}{1.mm} 
  \caption{Performance comparison of various heuristics for  16-buoy layouts for  the simplified irregular wave model (10 run each).}

  \label{tab:all}
  \scalebox{0.8}{
  \begin{tabular}{l|c|c|c|c|c|c|c|c|c|p{10cm}}
     \toprule
    \textbf{Methods}&\multicolumn{3}{c}{\textbf{PE-Full (Uniform)}} $|$& \multicolumn{3}{c}{\textbf{PE-Full(Normal)}}$|$& \multicolumn{3}{c}{\textbf{DE}} \\\hlineB{2}
&{$\mu=10$}&{$\mu=50$}&{$\mu=100$} 
&{$\mu=10$}&{$\mu=50$}&{$\mu=100$}                        
&{$P_{cr}=0.3$}&{$P_{cr}=0.5$}&{$Pcr=0.9$}   \\
    \midrule
    \texttt{Max} &6974948&6900024&6952017&6957388&6948746& 6892210& 7179681 & 7025873 & 7079962 \\
    \texttt{Median}&6859475&6839557&6851342& 6853987 & 6812866& 6816282 &6944795 & 6999356 & 6983523 \\
    \texttt{Mean}&6856337 &6821864&6860037& 6869586& 6837972& 6822553 & 6971231 & 6981195 & 6994172 \\
    \texttt{Std}&48701& 61880&50377 &61153 &70048 &52911 & 89244 & 41931 & 48943 \\
    \bottomrule
    \textbf{Methods} &\multicolumn{6}{ c }{\textbf{1+1EA}}$|$&  \multicolumn{2}{ c }{\textbf{ $NM_{2D}$} }$|$& \textbf{Iter-(1+1)EA} \\
    & {Mu-s=3} & {Mu-s=10} & {Mu-s=30} & {Linear} &{1/5 rule} & Fuzzy & {Uniform}& {Normal}\\
    \midrule
    \texttt{Max} & 7008380 & 7402584 & 7351112 & 7437481& 7425665&7454922 & 7380318& 7267242& 7370972\\
    \texttt{Median}& 6927230 & 7297465 & 7278120& 7317408& 7354589&7348676 & 7193110& 7136712& 7354589 \\
    \texttt{Mean}& 6908203 & 7292035 & 7275118& 7330286& 7343858&7335624 & 7205098& 7108693& 7274989 \\
    \texttt{Std}& 83157 & 77794& 51745& 60803 &59690&67061 & 83944& 116380&54380  \\
    \cline{1-10}
  \textbf{Methods} & \textbf{L$S_{1}$-N$M_{2D}$}&  \textbf{TDA} & \textbf{CMA-ES} & \textbf{R-S} &  \textbf{1+1E$A_{S}$}  & \textbf{(2+2)CMA-ES} &\textbf{L$S_{3}$-N$M_{2D}$} &\textbf{L$S_1$-N$M_{allDims}$} & \textbf{SLS}\\  
  \midrule
  \texttt  {Max}   &7608600  & 7148655 & 7118996 & 6825723 & 7370389 & 7205956& 7587758& 7094642& 7669439  \\
    \texttt{Median}&7409029  & 7057564 & 7053351 & 6658523 & 7214263 & 7073295 & 7459614  &6839911 &7590039\\
    \texttt{Mean}  &7427027 & 7005873 & 7038352 & 6676831 & 7236977 & 7080011& 7426742& 6823836& 7587410 \\
    \texttt{Std}   &129780 &  133977  & 84859   & 63883   & 67406   & 49771 & 123603& 198512 &52538\\
    \hlineB{2}
    \cline{1-10}
  \textbf{Methods} & \textbf{SLS+NM}&  \textbf{ISLS} & \textbf{ISLS-NM} & \textbf{ISL$S_{II}$-F} &  \textbf{ISL$S_{II}$-NM}  & \textbf{ISL$S_{II}$-SQP} &\textbf{ISL$S_{II}$-AS} &\textbf{IL$S_{3}$-N$M_{2D}$} &\textbf{ISL$S_{II}$-IP} \\ 
  \midrule
  \texttt  {Max}   &7680161 &7713744&7685633&7730799&7763249&7790679&\textbf{7878917} &7619404 & 7730844  \\
    \texttt{Median}&7548235&7647335 &7670401&7723825&7702161&7706788&7714011&7463928&7694381\\
    \texttt{Mean}  &7537582 &7651360&7666201 &7721954 & 7698639& 7701115&\textbf{7724370} &7469493&7688422  \\
    \texttt{Std}   &81727&25635&23437&9773&31790&60012&58341&102760&30453\\
    \hlineB{2}
  \end{tabular}
  }
\end{sidewaystable}
From these results, it can be seen that the most reliably performing placement algorithms are the variants of ISLS$_{(II)}$, with a run of ISLS$_{(II)}$-AS producing the best maximum performance with a maximum layout power of 7878917 Watts. In terms of statistical significance, the Active-Set, Nelder-Mead, and Fast-Search variants of ISLS$_{(II)}$  perform significantly better than the non ISLS$_{(II)}$ search methods with $p<0.025$ over the ten trial runs using a one-tailed Wilcoxon ranked-sum test. Moreover, the other ISLS$_{(II)}$ variants, SQP, and Interior-Point, perform better than all but the ISLS-NM search method. 

\begin{figure}[tbp]
\centering\includegraphics[width=0.8\textwidth]{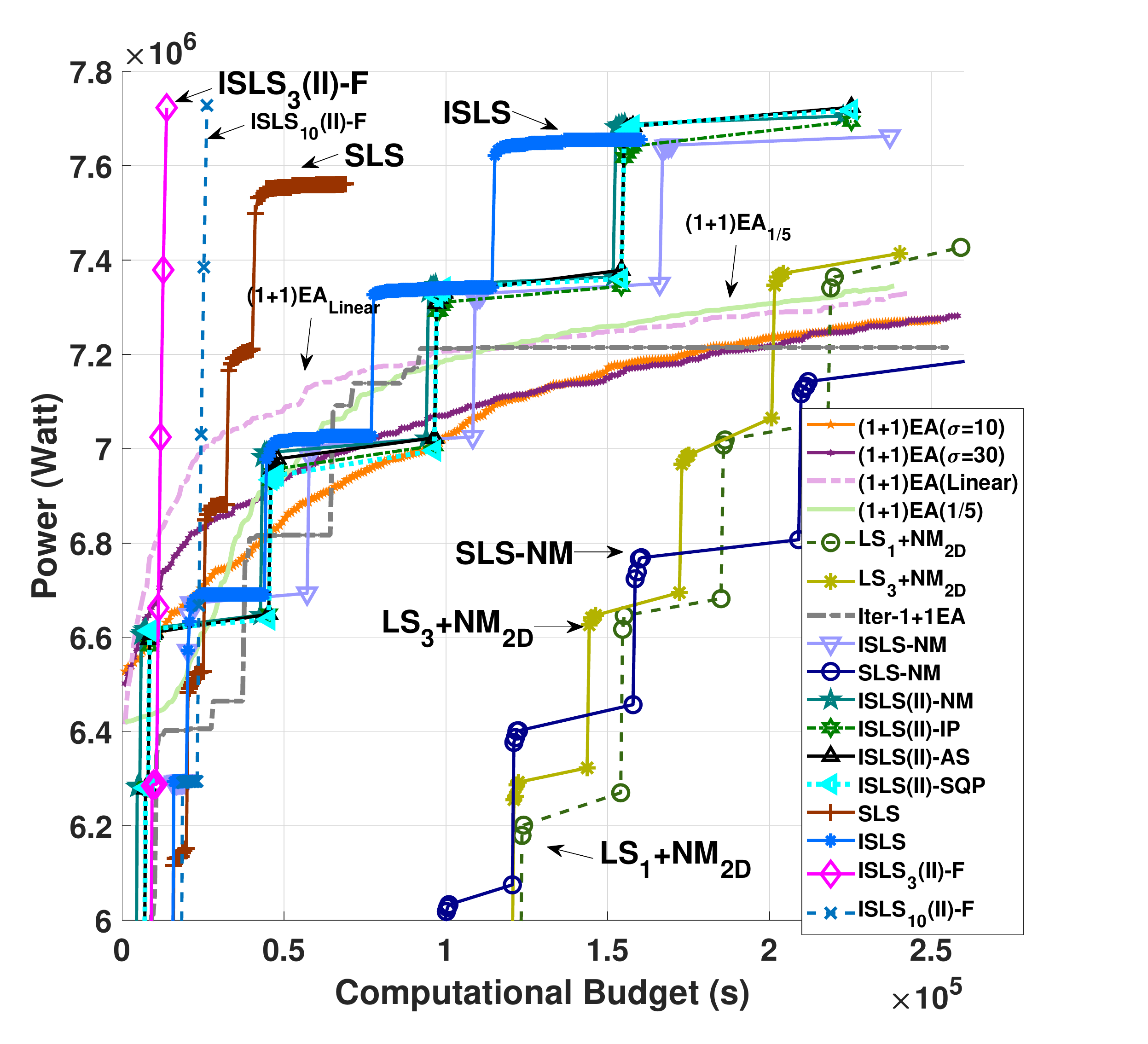}
\caption{The comparison of the average convergence rate of the proposed methods with the work \cite{neshat2018detailed} for 16-buoy layout over 72 hours (simplified irregular wave model).}
\label{fig:runtimeall16buoys}
\ \end{figure}

In Figure.~\ref{fig:runtimeall16buoys} each curve visualizes the evolution of the average power produced by the best individual layout for each method over three days of computational budget. The search parameters used by most algorithms are tuned to take advantage of most of the search budget. The exceptions to this are the plain SLS and ISLS algorithms where the local sampling converges early and the 3-sample and 10 sample versions of ISLS$_{(II)}$-F  which place the last buoys very quickly with a fast distance-based proxy function. From the plots, it can be seen that all the other variants of ISLS$_{(II)}$ converge to a very similar level of performance in a very similar time to each other. 
The average performance of all of  ISLS$_{(II)}$ variants is above all others. Overall, the best performing variant: ISLS$_{(II)}$-AS extracts 4\% more power than the previous best published algorithm: LS$_1$+NM$_{2D}$\ignore{ (Table \ref{table:all_new_results})}. This small difference in performance equates to approximately \$175,000 US dollars in extra annual avenue for the wave farm.
\footnote{Based on an annualized average wholesale price of US \$65 per MWh average for the Australian market in 2017/18.}\interfootnotelinepenalty=10000

\subsection{Real Wave Scenarios}
In this section, we first present a quick summary of the results for 4-buoy layouts before progressing to the much more challenging 16-buoy layouts. As mentioned previously for the 16 buoy-layout problems the full computational budget of 3 days on 12 CPU cores was used wherever possible. In order to make a fair comparison, an equivalent number of evaluations was used for the 4-buoy layout problem. 
To see how algorithm performance translates to real wave scenarios we ran ten trials each for eight selected search methods (DE, CMA-ES, LS$_{3}$NM$_{2D}$ \cite{neshat2018detailed}, IDE \cite{fang2018optimization}, bGA \cite{sharp2018wave}, 2+2CMA-ES \cite{wu2016fast}, ISLS$_{(II)}$-F and ISLS$_{(II)}$-AS) for 4 and 16-buoy layouts on the Sydney, Perth, Adelaide and Tasmania wave scenarios. 

\subsubsection{4-buoy layout results}

Table~\ref{Table:Summary_realwave_4buoy} summarises the results of the seven best-performing search methods on the 4-buoy layout problem in the four real wave scenarios. The output for the search method with the best maximum performance in each wave scenario is highlighted in bold. This table shows that the best results for each of the seven methods shown are within 1\% of each other in terms of raw performance. All methodologies are able to produce layouts with a q-factor close to one. It is also clear from these results that the global CMA-ES and IDE methods consistently have the best performance across all wave scenarios (except in the Sydney wave scenario) and the performance variance is quite small.

 \begin{table}
 \centering
\caption{Summary of the best  4-buoy layouts per-experiment (Power (Watt)) for the real wave scenarios .}
\scalebox{0.8}{
\begin{tabular}{|p{1.4cm}|p{1.2cm}|p{1.8cm}|p{1.9cm}|p{2cm}|p{2cm}|p{2.2cm}|p{2.2cm}|}
\hline
\hlineB{4}
\multicolumn{5}{ c }\textbf{Sydney wave model} \\ \hlineB{4}
\textbf{Methods} 
&\textbf{DE} 
& \textbf{CMA-ES}
& \textbf{L$S_3$-N$M_{2D}$\cite{neshat2018detailed} }
&\textbf{IDE~\cite{fang2018optimization}}
&\textbf{bGA~\cite{sharp2018wave}}
&\textbf{2+2CMA-ES~\cite{wu2016fast}}
& \textbf{ISL$S_{(II)}$-AS}
\\ \hline \hline
\textbf{Max} 
  &412667
  &412705
  &412294
  &412683
  &\textbf{413061}
  &412796
  &411291
   \\  \hline
  \textbf{Median} 
  &412557
  &412488
  &411069
  &412529
  &\textbf{413028}
  &412424
  &410094
   \\  \hline
\textbf{Mean} 
  &412580
  &412477
  &410839
  &412560
  &\textbf{413004}
  &412350
  &409376
    \\  \hline
  \textbf{Std} 
  &63
  &140
  &1184
  &74
  &54
  &395
  &1534
    \\  \hlineB{4}
  \multicolumn{5}{ c }\textbf{Perth wave model}  
   \\ \hline \hlineB{4}
 
\textbf{Max} 
  &398844
  &\textbf{399607} 
  &396759
  &\textbf{399607}
  &397822
  &399604
  &399476
   \\  \hline
  \textbf{Median} 
  &395898
  &\textbf{399607}
  &392753
  &\textbf{399607}
  &397822
  &399601
  &399466
   \\  \hline
\textbf{Mean} 
  &396615 
  &399117
  &391361
  &\textbf{399607}
  &397822
  &399600
  &399467
    \\  \hline
  \textbf{Std} 
  &1415
  &1033
  &5543
  &0.003
  &0.00
  &1.80
  &3.65
    \\  \hlineB{4} 
     \multicolumn{5}{ c }\textbf{Adelaide wave model}  
    \\ \hline \hlineB{4}

  \textbf{Max} 
  &399431
  &\textbf{402278}
  &401858
  &\textbf{402278}
  &402072
  &402276
  &402206
   \\  \hline
  \textbf{Median} 
  &397176 
  &\textbf{402278}
  &398352
  &\textbf{402278}
  &402072
  &402274
  &402186
   \\  \hline
\textbf{Mean} 
  &395620
  &402073
  &396106
  &\textbf{402278}
  &402072
  &402273
  &402189
    \\  \hline
  \textbf{Std} 
  &4271
  &709
  &6685
  &0.025
  &0.00
  &2.42
  &6.17
    \\  \hlineB{4}
    \multicolumn{5}{ c }\textbf{Tasmania wave model}  
    \\ \hline \hlineB{4}
  \textbf{Max} 
  &1093468
  &\textbf{1094611}
  &1094524
  &\textbf{1094611}
  &1072416
  &1094605
  &1094530
   \\  \hline
  \textbf{Median} 
  &1090833
  &\textbf{1094611}
  &1079619
  &\textbf{1094611}
  &1072379
  &1094596
  &1094524
   \\  \hline
\textbf{Mean} 
  &1090734 
  &\textbf{1094611}
  &1079429
  &\textbf{1094611}
  &1072190
  &1094597
  &1094523
    \\  \hline
  \textbf{Std} 
  &1985
  &0.0072
  &10432
  &0.008
  &471
  &4.078
  &5.50
    \\  \hlineB{4}
\end{tabular}
}
\label{Table:Summary_realwave_4buoy}
\end{table}
\begin{figure}[htbp!]
\centering
\includegraphics[width=0.6\textwidth]{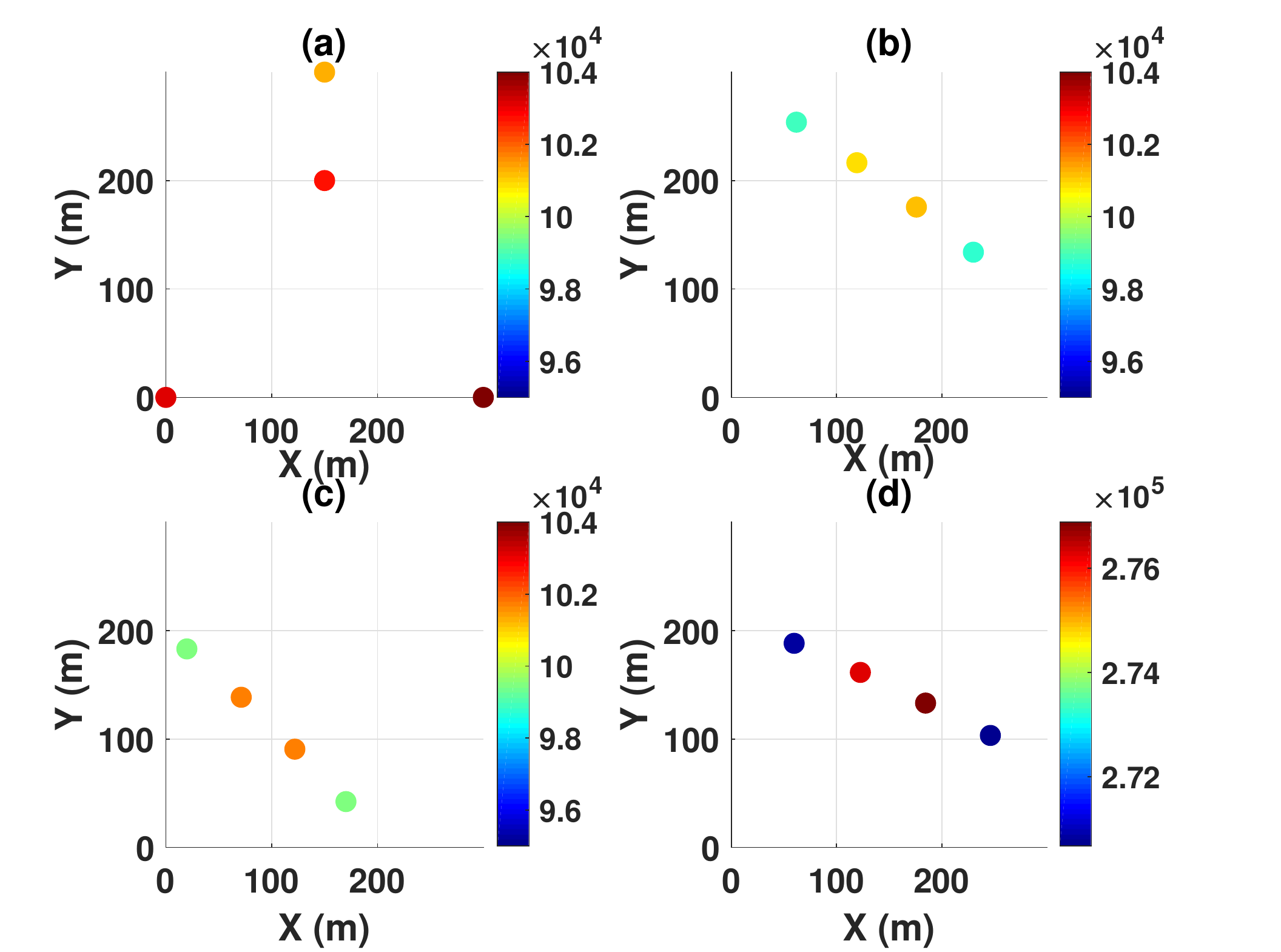}
\caption{ The best 4-buoy layouts of the real wave scenario by bGA (a) (Sydney:Power=413060 (Watt), q-factor=0.976 ), (b) (Perth: Power=399607 (Watt), q-factor=1.0366 ),  (c) (Adelaide: Power=402278 (Watt), q-factor=1.036) and  (d) (Tasmania: Power=1094611 (Watt), q-factor=1.0334). }
\label{fig:best_4buoy_realwave}
 \end{figure}
The 4-buoy layouts produced are shown in Figure.~\ref{fig:best_4buoy_realwave}. The buoys are coloured based on their power output. From these layouts, we can observe that, except for the Sydney wave scenario, all buoys form a row with the spacing and orientation determined by the wave environment. The orientation of this row (in the Perth, Adelaide and Tasmania scenarios) is aligned to the norm of the predominate wave direction for each scenario. We can also see that the middle two buoys in these three layouts also produce slightly more energy than the outer buoys. This is due to constructive interactions between buoys. The layout for the Sydney wave scenario differs markedly in its spacing and orientation. The Sydney wave environment is more varied in terms of wave direction and, thus, opportunities to exploit constructive interactions through a static layout are much reduced. As a result, the buoys in the produced layouts are rather spread out, indicating that the algorithms attempted to minimize destructive interactions. 

\subsubsection{16-buoy layout results}

\begin{figure}[bthp]
\centering
\subfloat[]{
\includegraphics[clip,width=0.49\columnwidth]{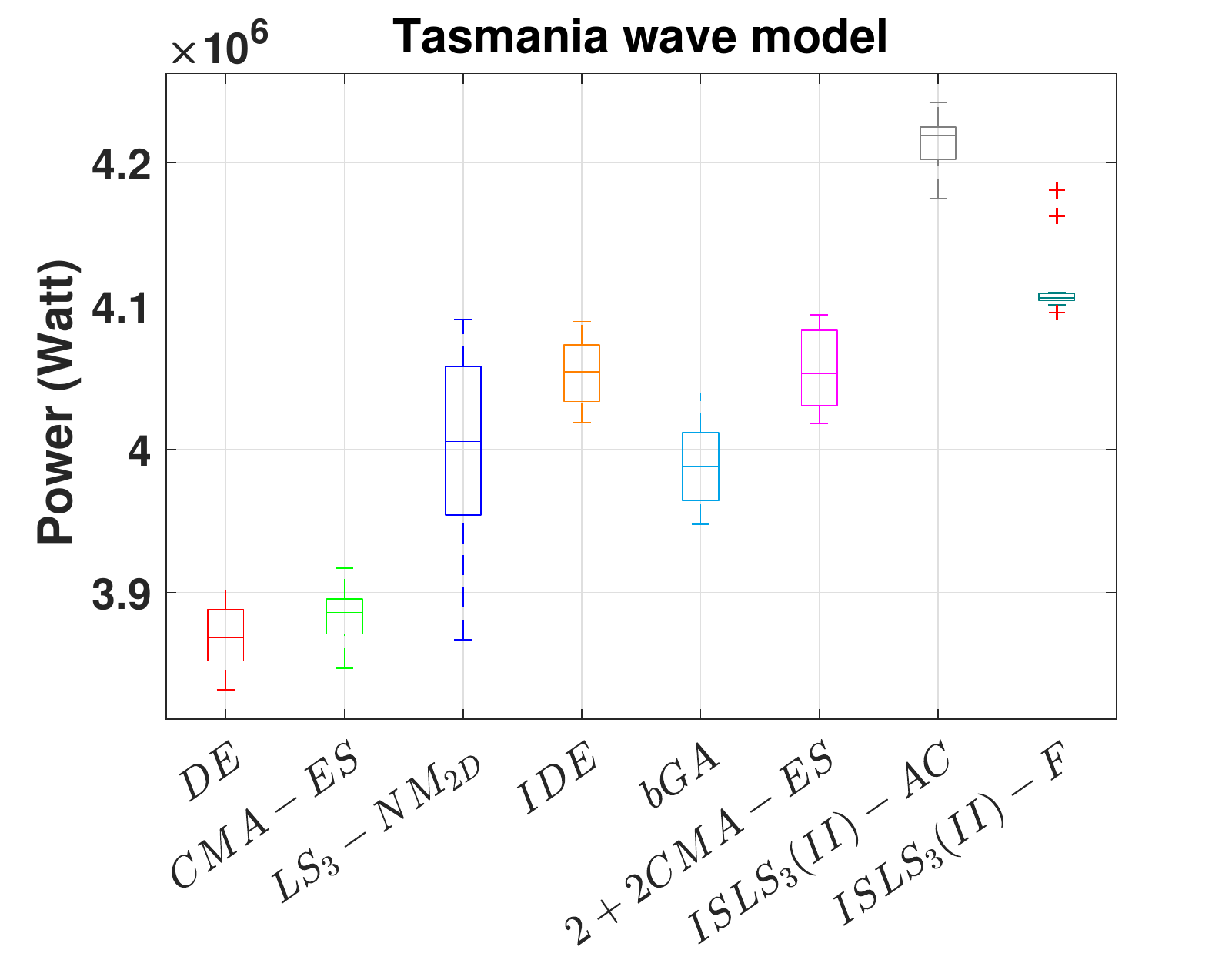}}
\subfloat[]{
\includegraphics[clip,width=0.49\columnwidth]{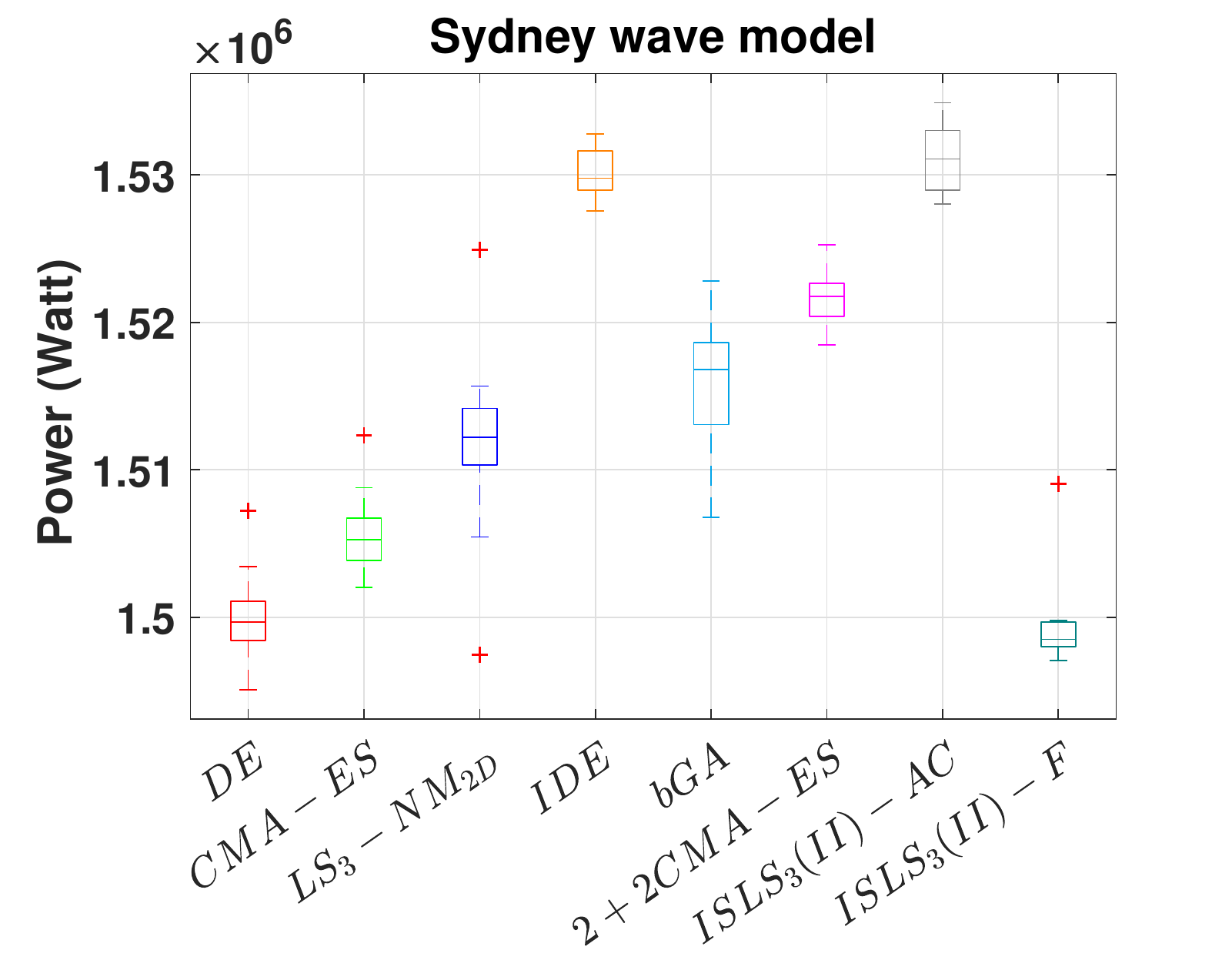}}\\
\subfloat[]{
\includegraphics[clip,width=0.49\columnwidth]{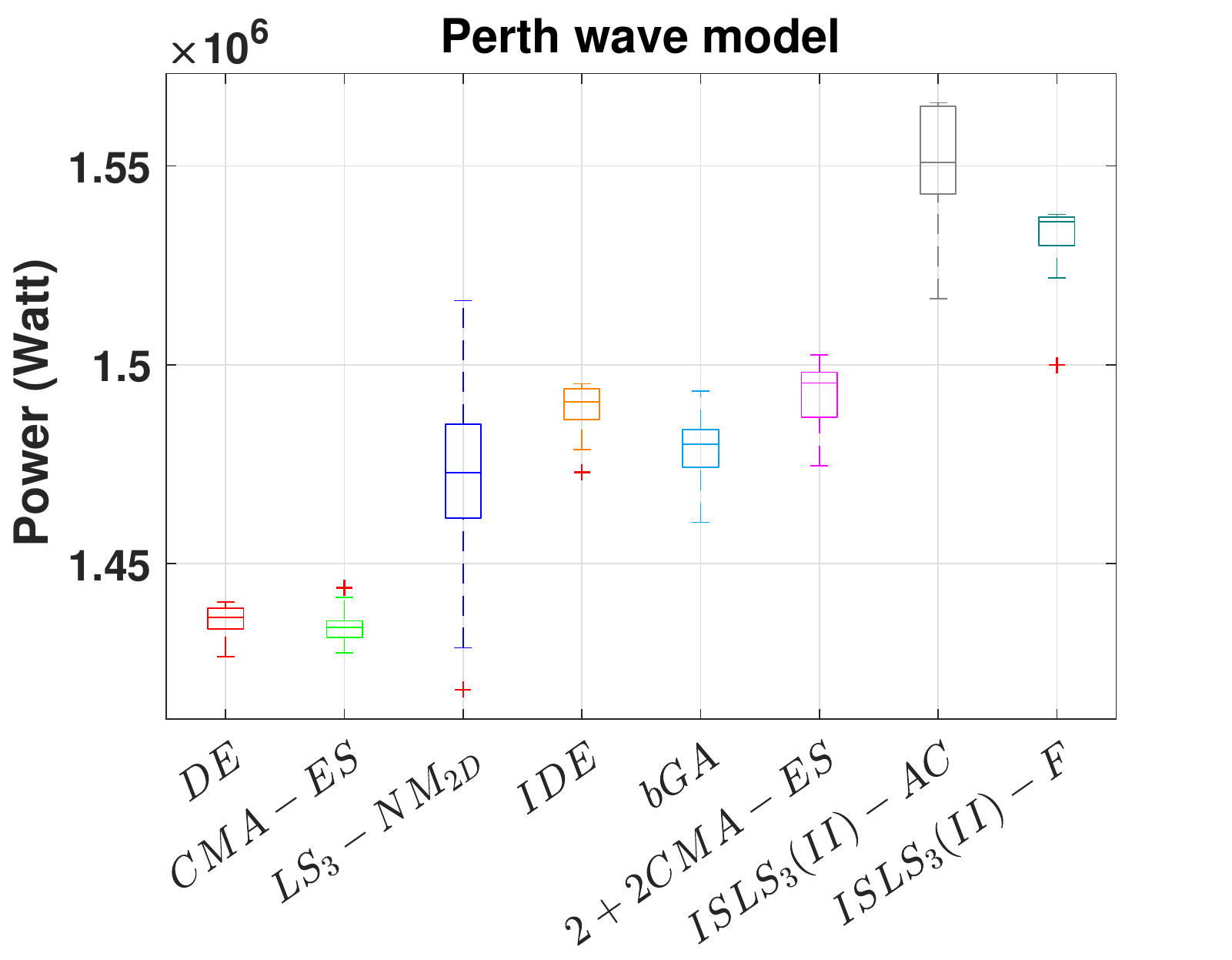}}
\subfloat[]{
\includegraphics[clip,width=0.49\columnwidth]{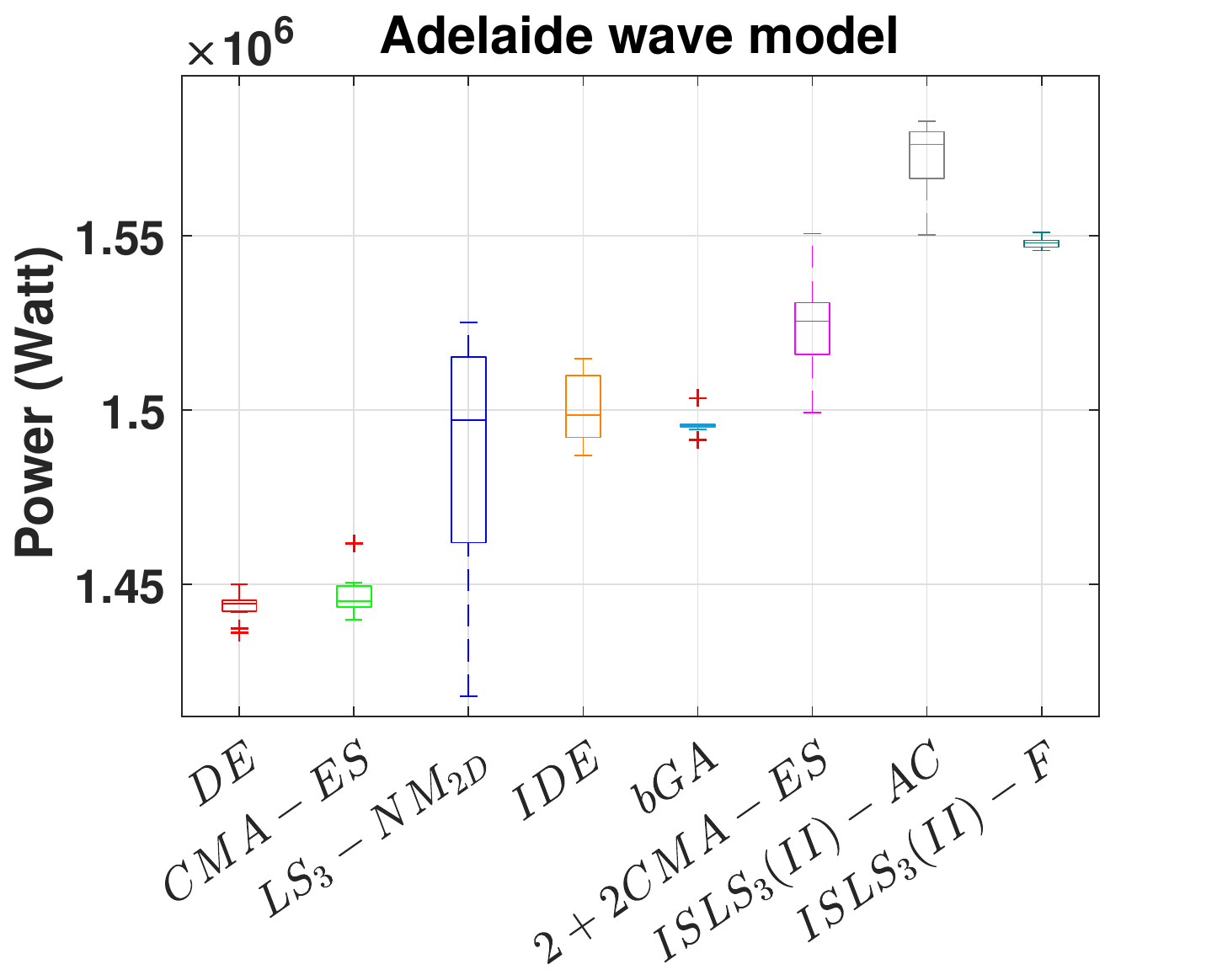}}
\caption{Optimization results of the proposed algorithms based on the best layout per run for 16 buoys and four real wave scenarios.((a) Tasmania wave scenario, (b) Sydney wave scenario , (c) Perth wave scenario , (d) Adelaide wave scenario)  }%
\label{fig:Summary_realwave}%
\end{figure}

Compared to the 4-buoy layout problem, the 16-buoy layout problem is challenging in terms of both problem constraints and computational budget. The farm area for the 16-buoy layout is larger than that for the 4-buoy layout, but not so large as to allow all 16 buoys to be placed in a single line in any wave scenario.
 The ISLS$_{(II)}$-AS performed significantly better than the other seven methods with $p<0.025$ over the ten trial runs using a one-tailed Wilcoxon ranked-sum test. Figure.~\ref{fig:Summary_realwave} summarises the results from these runs.
 
\begin{figure}[!tbhp]
\centering\includegraphics[width=0.75\textwidth]{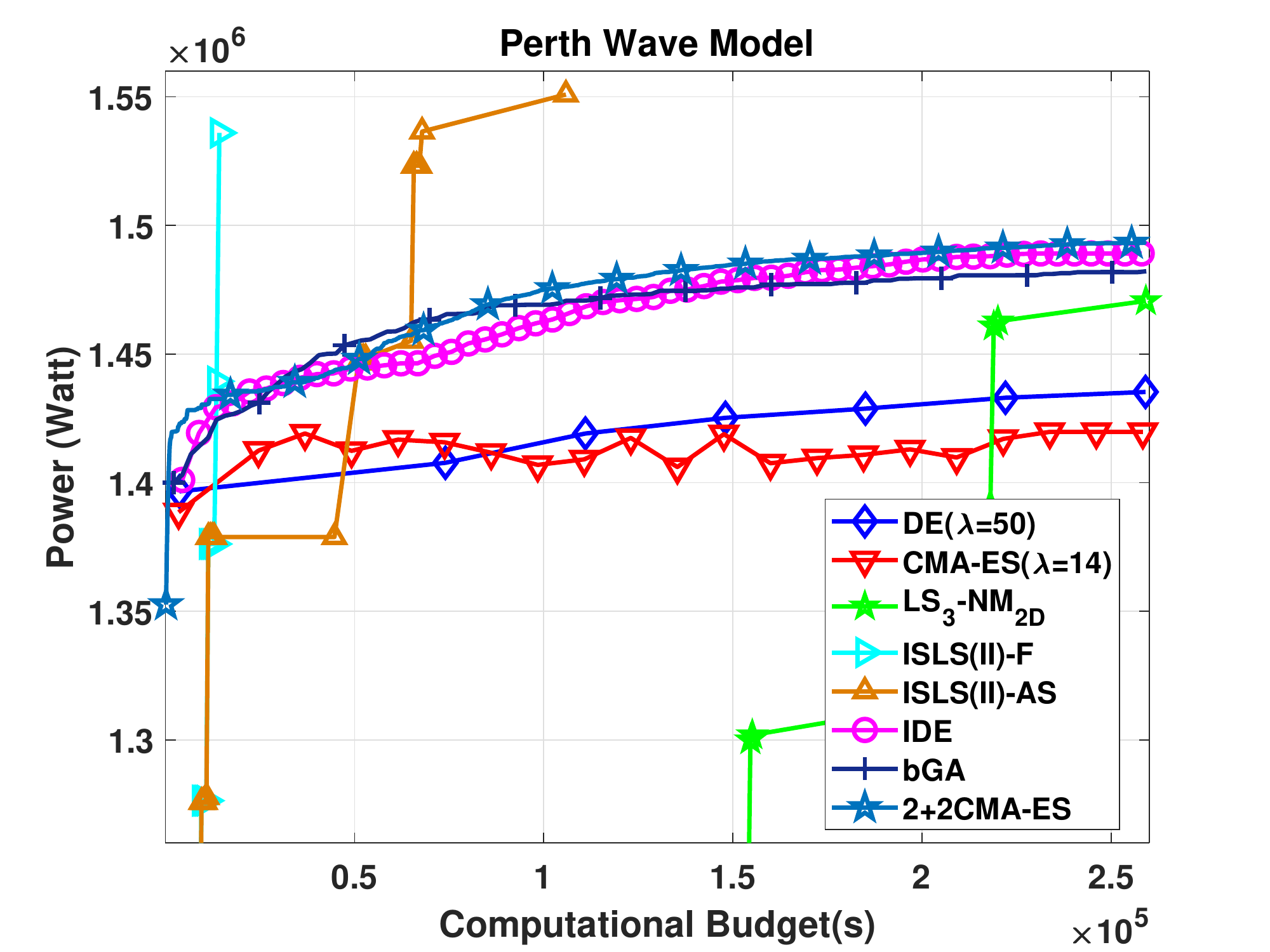}
\caption{The average convergence rate comparison of the proposed methods for N=16 in Perth wave model.
}
\label{fig:Perth16buoys}
 \end{figure}

These results are reflected in the clear margin between the performance of ISLS$_{(II)}$-AS and the other methods. It can be seen that ISLS$_{(II)}$-F varies significantly in its performance between scenarios with much poorer performance for Sydney. This perhaps reflects on diminished usefulness for the distance-based proxy function in this complex wave environment. To sum up, in all cases, ISL$S_{(II)}$-AS provides the best mean and maximal power output among the eight compared algorithms statistically. A comparison of the ISLS$_{(II)}$-AS convergence rate with seven other methods for Perth wave model can be seen in Figure \ref{fig:Perth16buoys}. The figure illustrates that ISLS$_{(II)}$-F has the highest convergence speed; however, the average quality of proposed 16-buoy layouts of ISLS$_{(II)}$-AS can be considerably better.

\begin{figure}[htbp!]
\centering
\includegraphics[width=1.1\textwidth]{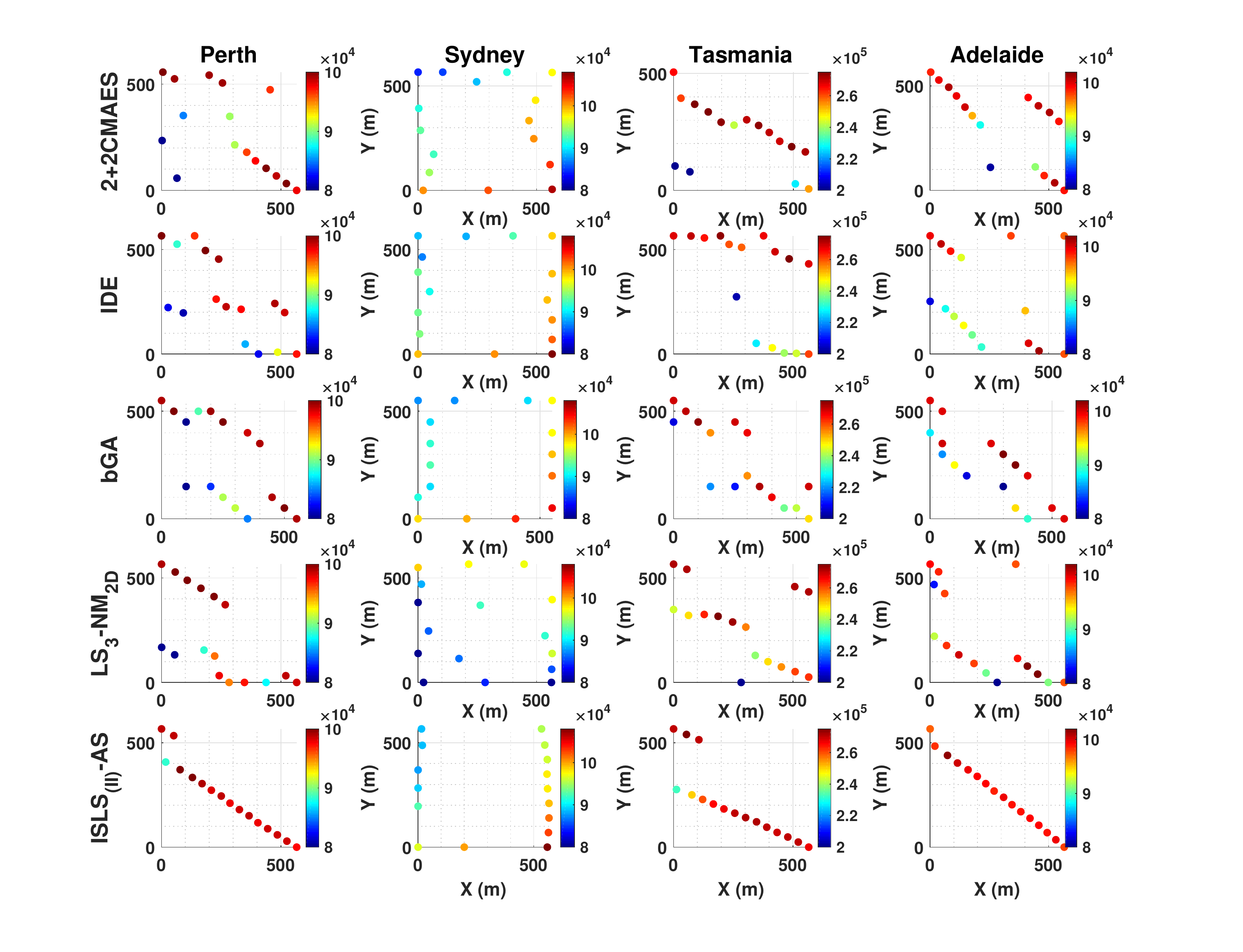}
\caption{ The best-found 16-buoy layouts of the real wave scenarios by ISL$S_{(II)}$-AS::Sydney: Power=1,534.9~kW, q-factor=0.9068,  Perth: Power=1,565.6~kW, q-factor=1.015, Adelaide: Power=1,583.1~kW, q-factor=1.019 and Tasmania: Power=4,241.8~kW, q-factor=1.0012. The absorbed power of other 16-buoy layout can be seen in Table \ref{Table:Summary_realwave}.}
\label{fig:best_16buoy_realwave}
 \end{figure}

The best 16-buoy layouts produced by ISLS$_{(II)}$-AS in each of the four wave scenarios are shown in Figure.~\ref{fig:best_16buoy_realwave}. For Perth, Adelaide and Tasmania the layouts are similar with the buoys being oriented in a line roughly normal to the prevailing wave direction with buoys placed in phase 1 (lower numbers) forming a bottom row and the buoys placed in phase 2 (higher numbers) behind these. It can be seen that fewer buoys are placed in phase 1 for Tasmania than for Adelaide. The number of buoys placed in this phase appears to depend on how well the first row is aligned with the diagonal of the farm area. 

\begin{table}%
\centering%
\caption{Summary of the best achieved 16-buoy layouts power(Watt) per experiment for real wave scenarios }
\scalebox{0.8}{
\begin{tabular}{|p{1.4cm}|p{1.2cm}|p{1.7cm}|p{1.5cm}|p{1.5cm}|p{1.5cm}|p{1.7cm}|p{1.8cm}|p{2.1cm}|}
\hline
\hlineB{4}
\multicolumn{6}{ c }\textbf{Sydney wave model} \\ \hlineB{4}
\textbf{Methods} 
&\textbf{DE} 
& \textbf{CMA-ES}
& \textbf{L$S_3$-N$M_{2D}$\cite{neshat2018detailed} }
&\textbf{IDE~\cite{fang2018optimization}}
&\textbf{bGA~\cite{sharp2018wave}}
&\textbf{2+2CMA-ES~\cite{wu2016fast}}
&\textbf{ISL$S_{(II)}$-F}
& \textbf{ISL$S_{(II)}$-AS}
\\ \hline \hline
\textbf{Max} 
  &1507235
  &1512337
  &1524915
  &1532776
  &1522817
  &1525243
  &1509037
  &\textbf{1534883}
   \\  \hline
  \textbf{Median} 
  &1499663
  &1505254
  &1512190
  &1529766
  &1516803
  &1521448
  &1498963
  &\textbf{1531785}
   \\  \hline
\textbf{Mean} 
  &1499901
  &1505705
  &1511814
  &1530250
  &1516061
  &1521451
  &1499663
  &\textbf{1531491}
    \\  \hline
  \textbf{Std} 
  &3207
  &2770
  &6421
  &1961
  &4617
  &1972
  &3440
  &2339
    \\  \hlineB{4}
   \multicolumn{6}{ c }\textbf{Perth wave model}  
 
   \\ \hline \hlineB{4}
\textbf{Max} 
  &1440344
  &1443893 
  &1516098
  &1495206
  &1493394
  &1502466
  &1537788
  &\textbf{1565836}
   \\  \hline
  \textbf{Median} 
  &1436441
  &1433949
  &1472851
  &1489437
  &1479940
  &1494378
  &1529076
  &\textbf{1550877}
   \\  \hline
\textbf{Mean} 
  &1435539 
  &1434333
  &1470658
  &1488187
  &1479265
  &1491837
  &1528225
  &\textbf{1549409}
    \\  \hline
  \textbf{Std} 
  &4171
  &4626
  &26990
  &6642
  &8991
  &8713
  &9016
  &16920
    \\  \hlineB{4} 
     \multicolumn{6}{ c }\textbf{Adelaide wave model} 
    \\ \hline \hlineB{4}
\textbf{Max} 
  &1449967
  &1461741
  &1525144
  &1514816
  &1511594
  &1550701
  &1551102
  &\textbf{1583052}
   \\  \hline
  \textbf{Median} 
  &1444455 
  &1445241
  &1497211
  &1496515
  &1492807
  &1527727
  &1547351
  &\textbf{1578797}
   \\  \hline
\textbf{Mean} 
  &1443442
  &1446621
  &1477342
  &1500835
  &1493705
  &1525274
  &1548027
  &\textbf{1573476}
    \\  \hline
  \textbf{Std} 
  &3757
  &5879
  &57675
  &10207
  &10190
  &13695
  &1796
  &11694
    \\  \hlineB{4}
    \multicolumn{6}{ c }\textbf{Tasmania wave model}  
    \\ \hline \hlineB{4}
\textbf{Max} 
  &3901664
  &3916983
  &4090733
  &4089215
  &4050476
  &4093637
  &4180781
  &\textbf{4241838}
   \\  \hline
  \textbf{Median} 
  &3868558
  &3886093
  &4005319
  &4052658
  &4008556
  &4066558
  &4105700
  &\textbf{4218894}
   \\  \hline
\textbf{Mean} 
  &3867923 
  &3882930
  &3999507
  &4053854
  &4002820
  &4058384
  &4115758
  &\textbf{4213652}
    \\  \hline
  \textbf{Std} 
  &22588
  &19705
  &69758
  &25983
  &27647
  &29421
  &26696
  &21775
    \\  \hlineB{4}
\end{tabular}
}
\label{Table:Summary_realwave}

\end{table}

Again, the Sydney layout is very different, with a row for phase 1 oriented to the east and the other buoys being placed at large distances from the others. We have observed this pattern for a number of Sydney runs where the best layouts tended to contain widely dispersed buoys to minimize destructive interference. 

One interesting observation from the layouts in Figure.~\ref{fig:best_16buoy_realwave}\ignore{\ref{fig:wave_interpolation}} is that there are buoys on the end of the front row for both Adelaide and Tasmania which produce less energy than the buoys behind them. While this appears to be counter-intuitive at first, it is simply the result of complex interactions between the front row of buoys and the ocean waves. 

Figure.~\ref{fig:wave_interpolation} demonstrates the nature of these interactions by showing the power landscape for the Perth wave scenario for four different layouts. In all cases, the wave energy in the area in the lee of the front row of the buoys (top right) is {\em{greater}} than the area immediately in front. This phenomenon is due to the fact that buoys interact strongly with the surrounding water on all sides. In fact, modelling over an extended area shows that the wave-damping influence of these layouts stretches more than 500 meters further out to sea than the buoy array. This extended impact helps illustrate the potential for even relatively small buoy arrays to extract energy from a relatively large area of ocean. 
Another observation is of the relative efficacy of ISLS(II)-AS which places all of its buoys in high-energy locations. 
\begin{figure}[!tb]
\centering\includegraphics[width=0.6\textwidth]{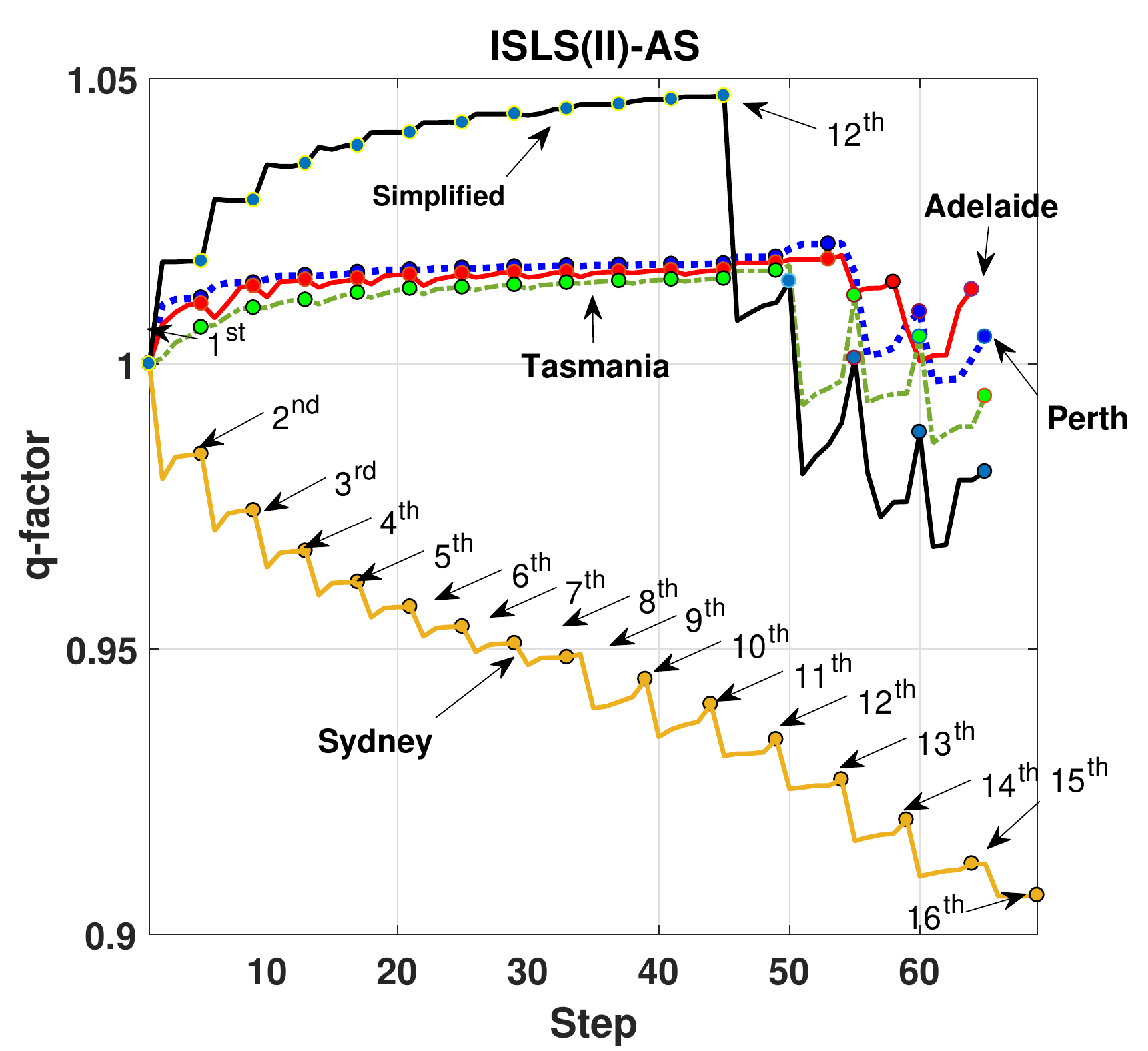}
  \caption{The average of ISLS(II)-AS q-factor  performance (16-buoy) for real wave models and the simplified irregular model.}
  \label{fig:ISLS_AS_q_factor}

\end{figure}

The impact for constructive and destructive interference in each environment can be visualized in the trajectory of average q-factor as ISLS$_{(II)}$-AS places buoys in each of the wave scenarios. Figure~\ref{fig:ISLS_AS_q_factor} shows this trajectory for the four wave scenarios and, as a reference to the simplified irregular wave scenario. It can be seen that Adelaide, Perth, and Tasmania are characterized by constructive interference in phase 1 and both Adelaide and Perth still have a q-factor greater than one even after phase 2. In contrast, the Sydney wave scenario is characterized by destructive interference throughout the search -- though still producing net gains in power output for each buoy placed. Our code and auxiliary material are publicly available: \url{https://cs.adelaide.edu.au/~optlog/research/energy.php}

 \begin{figure}[!tb]
 \centering
\includegraphics[width=0.8\textwidth]{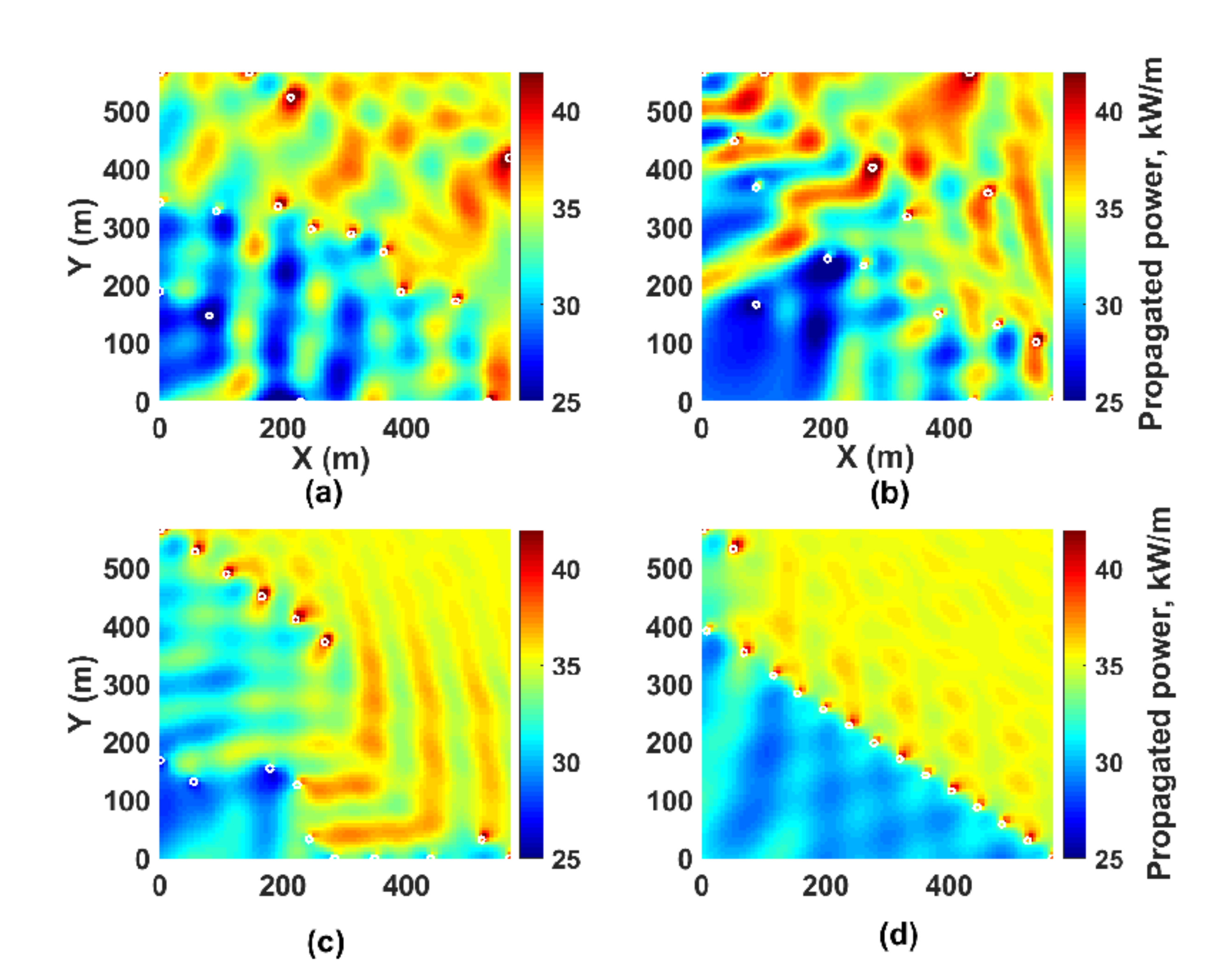}
\caption{Interpolated wave energy landscape for the best 16-buoy layouts for Perth wave scenario, a) CMA-ES, b) DE, c) LS-NM and d) ISLS(II)-AS. White circles represent the buoy placement. (the wave angle propagates at 232.5 degrees). }
\label{fig:wave_interpolation}

 \end{figure}

\section{Conclusion}
In this work, we have explored algorithmic solutions to the problem of placing wave-energy converter buoys in arrays in order to maximize energy output. We have developed, evaluated, and systematically compared nine new search heuristics to a range of existing standard and domain-specific search techniques. The algorithms were benchmarked on both artificial and real wave scenarios. Producing effective layouts presented interesting challenges in terms of the high cost of full function evaluations (approximately 700 seconds for one evaluation of a 16-buoy farm), but also complex problem dynamics including multi-modal constructive and destructive interference between buoys and highly varied power landscapes for buoy placement. 
In this work, the algorithms that performed best in experiments were hybrid search heuristics that used local search informed by inexpensive proxy models that were customized for local conditions. These methods further optimized the cost of function evaluations by placing one buoy at a time -- thus minimizing the number of modeled interactions. The most effective search techniques of all used the proxy model to inform the placement of the first row of buoys and switched to a combination of local search and gradient search techniques once the farm boundary was reached. 

One possible limitation of the best approaches described here is that they allow no backtracking to further optimize buoy positions once they have been placed. Preliminary experiments with further global optimization of four buoy layouts have shown some small potential gains from further global optimization, though at a much much greater (87-fold) computational cost.
This work can be extended in several ways. First, while the hydrodynamic models employed here are state-of-the-art in terms of fidelity, the wave farm environments are still simplified in terms of farm geometry (assumed to be squared) and seafloor topography (assumed to be uniform in depth). In the future, both of these assumptions can be relaxed with farm geometry allowed to vary to match realistic lease-boundaries and the hydrodynamic model updated to allow varied seafloor depth. It is also possible to increase the complexity of the model for each buoy in terms of size, depth, and tether parameters. Some of these parameters impact on cost and, thus, produce scope for multi-objective optimization. Finally, there is scope to learn a robust and accurate proxy function for evaluating energy outputs. Such an approach might use machine learning techniques such as Deep Neural Networks to act as a partial or complete estimator function for the output of a given layout. Such an estimator has the potential to greatly increase the speed of search and open the way for further improvements in search heuristics.

\bibliographystyle{unsrt}  
\bibliography{sample-bibliography}

\end{document}